\documentclass[sigconf,nonacm]{acmart}
\AtBeginDocument{%
  \providecommand\BibTeX{{%
    \normalfont B\kern-0.5em{\scshape i\kern-0.25em b}\kern-0.8em\TeX}}}

\setcopyright{acmcopyright}
\copyrightyear{2026}
\acmYear{2026}
\usepackage{graphicx}
\usepackage{float} 
\usepackage{subcaption}
\usepackage{utfsym}
\usepackage{multirow}
\usepackage{enumitem}
\usepackage{mdframed}
\usepackage{amsmath, amssymb}
\usepackage{makecell}
\usepackage{pifont}

\usepackage{multirow}
\usepackage{balance}
\usepackage{adjustbox} 
\usepackage{threeparttable}
\usepackage{xcolor}
\usepackage{color,colortbl}
\usepackage{tikz}
\usepackage{booktabs}
\usepackage{tikz-qtree}
\usetikzlibrary{fit,calc}
\usepackage{pgfplots}
\usepackage{varwidth}
\usepackage{url}
\usepackage{mathtools}
\usepackage{rotating} 

\usepackage{amsmath}
\usepackage{graphicx}
\usepackage{todonotes}
\usepackage{cleveref}

\usepackage[ruled,linesnumbered,vlined]{algorithm2e}
\usetikzlibrary{patterns}

\usepackage[noend]{algpseudocode}

\newcommand{\revise}[1]{#1}

\definecolor{linecolor}{rgb}{0.55, 0, 0} 
\definecolor{bgcolor}{rgb}{1, 0.95, 0.95} 

\usepackage[most]{tcolorbox}

\usepackage{tabularx}

\tcbset{highlightcode/.style=
    {enhanced, 
    colframe=gray!20, 
    colback=gray!20, 
    boxrule=0pt, 
    boxsep=0pt, 
    left=1mm, 
    right=0mm, 
    top=1mm, 
    bottom=1mm,
    overlay={
        \begin{scope}[shift={(frame.north west)}]
            \foreach \i in {1,2,3} {
                \node[anchor=east] at (-5mm,{-\i*15mm + 1.5mm}) {\footnotesize\the\numexpr\value{AlgoLine}+\i\relax};
            }
        \end{scope}
    },
}}
\usepackage{framed}

\definecolor{mycolor1}{RGB}{53,122,162}   
\definecolor{mycolor2}{RGB}{164,224,187} 

\newcommand{\colorcell}[1]{%
    \cellcolor{mycolor1!#1!mycolor2}%
    \ifnum#1>60
      {\Large \textcolor{white}{#1}}%
    \else
      {\Large \textcolor{black}{#1}}%
    \fi
}

\newcommand\vldbdoi{XX.XX/XXX.XX}
\newcommand\vldbpages{XXX-XXX}
\newcommand\vldbvolume{19}
\newcommand\vldbissue{1}
\newcommand\vldbyear{2026}
\newcommand\vldbauthors{\authors}
\newcommand\vldbtitle{\shorttitle} 
\newcommand\vldbpagestyle{plain} 

\newcommand{\cmark}{\ding{51}} 
\newcommand{\xmark}{\ding{55}} 

\usepackage[most]{tcolorbox}
\usepackage{float}
\usepackage{xspace}
\tcbset{
  aibox/.style={
    width=\linewidth,
    top=6pt,
    bottom=2pt,          
    left=2pt,            
    right=2pt,           
    colback=white,
    colframe=black,
    colbacktitle=black,
    enhanced,
    center,
    attach boxed title to top left={yshift=-0.1in,xshift=0.15in},
    boxed title style={boxrule=0pt,colframe=white,},
  }
}
\newtcolorbox{AIbox}[2][]{aibox,title=#2,#1}

\begin{document}
\title{Memory in the LLM Era: Modular Architectures and Strategies in a Unified Framework [Experiment, Analysis \& Benchmark]}







\author{Yanchen Wu}
\authornote{The first two authors contributed equally to this research.}
\affiliation{
  \institution{CUHK-Shenzhen}
  \city{}
  \country{}
}

\author{Tenghui Lin}
\authornotemark[1]
\affiliation{
  \institution{CUHK}
  \city{}
  \country{}
}

\author{Yingli Zhou}
\authornote{These authors jointly led this project.}
\affiliation{
  \institution{CUHK-Shenzhen}
  \city{}
  \country{}
}

\author{Fangyuan Zhang}
\authornotemark[2]
\affiliation{
  \institution{HITSZ}
  \city{}
  \country{}
}

\author{Qintian Guo}
\affiliation{
  \institution{BIT}
  \city{}
  \country{}
}

\author{Xun Zhou}
\affiliation{
  \institution{HITSZ}
  \city{}
  \country{}
}

\author{Sibo Wang}
\affiliation{
  \institution{CUHK}
  \city{}
  \country{}
}

\author{Xilin Liu}
\affiliation{
  \institution{Huawei Cloud}
  \city{}
  \country{}
}

\author{Yuchi Ma}
\affiliation{
  \institution{Huawei Cloud}
  \city{}
  \country{}
}

\author{Yixiang Fang}
\affiliation{
  \institution{CUHK-Shenzhen}
  \city{}
  \country{}
}












\begin{abstract}
Memory emerges as the core module in the large language model (LLM)-based agents for long-horizon complex tasks (e.g., multi-turn dialogue, game playing, scientific discovery), where memory can enable knowledge accumulation, iterative reasoning and self-evolution.
A number of memory methods have been proposed in the literature. 
However, these methods have not been systematically and comprehensively compared under the same experimental settings.
%
\revise{In this paper, we first summarize a unified framework that covers existing representative agent memory methods from a high-level perspective.}
%
\revise{
We then extensively compare representative agent memory methods on two long-term conversational benchmarks and an agentic memory benchmark, and examine the effectiveness of representative methods, providing a thorough analysis of those methods.
}
As a byproduct of our experimental analysis, we also design a new memory method by exploiting modules in the existing methods, which outperforms the state-of-the-art methods.
Finally, based on these findings, we offer promising future research opportunities.
We believe that a deeper understanding of the behavior of existing methods can provide valuable new insights for future research.
\end{abstract}



\maketitle

\pagestyle{\vldbpagestyle}
\begingroup\small\noindent\raggedright\textbf{PVLDB Reference Format:}\\
\vldbauthors. \vldbtitle. PVLDB, \vldbvolume(\vldbissue): \vldbpages, \vldbyear.\\
\href{https://doi.org/\vldbdoi}{doi:\vldbdoi}
\endgroup
\begingroup
\renewcommand\thefootnote{}\footnote{\noindent
This work is licensed under the Creative Commons BY-NC-ND 4.0 International License. Visit \url{https://creativecommons.org/licenses/by-nc-nd/4.0/} to view a copy of this license. For any use beyond those covered by this license, obtain permission by emailing \href{mailto:info@vldb.org}{info@vldb.org}. Copyright is held by the owner/author(s). Publication rights licensed to the VLDB Endowment. \\
\raggedright Proceedings of the VLDB Endowment, Vol. \vldbvolume, No. \vldbissue\ %
ISSN 2150-8097. \\
\href{https://doi.org/\vldbdoi}{doi:\vldbdoi} \\
}\addtocounter{footnote}{-1}\endgroup

\ifdefempty{\vldbavailabilityurl}{}{
\vspace{.3cm}
\begingroup\small\noindent\raggedright\textbf{PVLDB Artifact Availability:}\\
The source code, data, and/or other artifacts have been made available at \url{https://github.com/Yanchen398/Memory-in-the-LLM-Era}.
\endgroup
}


\definecolor{1000w}{HTML}{AA0030} 

\definecolor{1w}{HTML}{6B238E}   
\definecolor{10w}{HTML}{00AA00}  

\definecolor{DarkCoral}{HTML}{CD5B45}
\definecolor{DarkSoftApricot}{HTML}{D1A655}

\newcommand{\zhou}[1]{\textcolor{magenta}{[zhou: #1]}}
\section{Introduction}
\label{sec:intro}

\begin{figure}[]
    \centering
    \setlength{\abovecaptionskip}{0cm}
    \setlength{\belowcaptionskip}{-0.3cm}
    \includegraphics[width=0.8\linewidth]{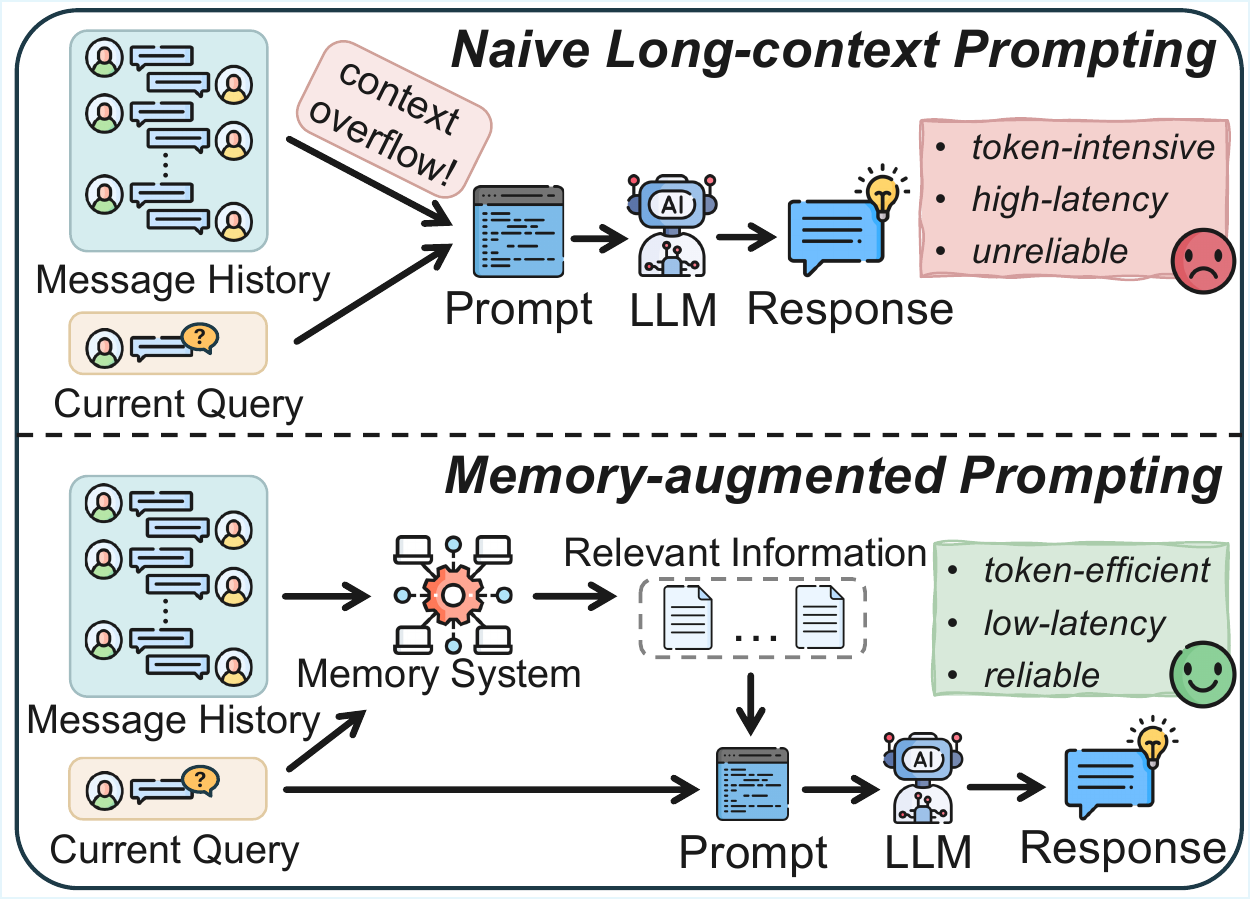}
    \caption{Overview of naive long-context prompting and memory-augmented prompting.}
    \label{fig:intro}
\end{figure}

\begin{table*}[]
  \centering
  \caption{\revise{Classification of representative agent memory methods.}}
  \renewcommand{\arraystretch}{1.25}
  \label{tab:image_memory_methods}
  \resizebox{\textwidth}{!}{
  \revise{
  \begin{tabular}{l|l|l|l|l}
     \toprule
     \rowcolor{gray!20}
      \textbf{Method} & \textbf{Information Extraction} & \textbf{Management Operations} & \textbf{Storage Structure} & \textbf{Retrieval Mechanism} \\
     \midrule
      {\tt A-MEM}~\cite{wujiang2025amem} & Direct archive, Summarization-based extract & Connect, Update & Flat, Vector & Vector-Based \\
      {\tt MemoryBank}~\cite{zhong2024memorybank} & Direct archive & Integrate, Update, Filter & Flat, Vector & Vector-Based \\
      {\tt MemGPT}~\cite{packer2023memgpt} & Direct archive & Integrate, Transform, Update & Hierarchical, Vector & Lexical-Based, Vector-Based \\
      {\tt Mem0}~\cite{chhikara2025mem0} & Direct archive, Summarization-based extract & Integrate, Update, Filter & Flat, Vector & Vector-Based \\
      {\tt Mem0$^g$}~\cite{chhikara2025mem0} & Graph-based extract & Connect, Update, Filter & Flat, Graph & Vector-Based, Structure-Based \\
      {\tt MemoChat}~\cite{lu2023memochat} & Direct archive & Integrate & Flat & LLM-Assisted \\
      {\tt Zep}~\cite{rasmussen2025zep} & Direct archive, Graph-based extract & Connect, Transform, Update & Hierarchical, Graph & Lexical-Based, Vector-Based, Structure-Based \\
      {\tt MemTree}~\cite{rezazadeh2024isolated} & Direct archive & Connect, Integrate, Update & Flat, Tree & Vector-Based \\
      {\tt MemoryOS}~\cite{kang2025memory} & Direct archive & Connect, Integrate, Transform, Update, Filter & Hierarchical, Vector & Lexical-Based, Vector-Based \\
      {\tt MemOS}~\cite{li2025memos} & Direct archive, Summarization-based extract & Connect, Integrate, Update & Hierarchical, Tree & Lexical-Based, Vector-Based \\
      {\tt MemGAS}~\cite{xu2026memgas} & Direct archive, Summarization-based extract & Connect & Flat, Graph & Vector-Based, Structure-Based, LLM-Assisted \\
      {\tt LightMem}~\cite{fang2025lightmem} & Direct archive, Summarization-based extract & Integrate, Update, Filter & Hierarchical, Vector & Vector-Based \\
     \bottomrule
  \end{tabular}
  }
  }
\end{table*}

The development of Large Language Models (LLMs) like GPT-5~\cite{singh2025openaigpt5card}, Qwen3~\cite{yang2025qwen3}, and Claude Sonnet 4.6~\cite{anthropic2026claude46} has sparked a revolution in the field of artificial intelligence~\cite{liu2024survey,huang2023survey,wang2024survey,zheng2024large,li2023large,nie2024survey,ghimire2024generative,wang2024large}.
Building on this success, LLM-powered agents have rapidly emerged and are being deployed across a wide range of domains, from industrial automation to personal assistance. 
For example, SWE-agent systems for software engineering tasks~\cite{yang2024sweagentagentcomputerinterfacesenable} and personal assistant agents such as OpenClaw~\cite{openclaw2026} illustrate how LLM-based agents can autonomously plan, reason, and execute complex multi-step workflows.
These agents are increasingly expected to operate autonomously, adapt to diverse environments, and support personalized interactions tailored to user needs.
A key capability underlying such intelligent behavior is the {\it memory mechanism}~\cite{zhang2018personalizingdialogueagentsi,park2023generativeagents}. 
As illustrated in Figure~\ref{fig:intro}, memory mechanisms allow agents to move beyond naive long-context prompting by maintaining and leveraging relevant information from past interactions.
By equipping agents with memory mechanisms, they can accumulate experience over time, maintain contextual knowledge, and make more informed decisions—analogous to how humans rely on memory to learn from past experiences and guide future actions.

In recent years, a growing number of memory methods have been proposed to enhance the capability of LLM-based agents to retain, organize, and utilize historical information across interactions. These methods aim to enable agents to move beyond stateless reasoning and instead support long-term planning, personalization, and adaptive decision-making.
Motivated by the limitations of stateless LLM agents and the growing need for persistent contextual reasoning, researchers from multiple communities---including databases, data mining, machine learning, and natural language processing---have begun developing efficient and scalable memory mechanisms for intelligent agents~\cite{rezazadeh2024isolated, packer2023memgpt, park2023generativeagents, weng2023agent, wang2024llmagentsurvey, zhou2025mem1, ouyang2025reasoningbank, kagaya2024rap, wang2024awm, wang2025m+}.
\revise{In Table~\ref{tab:image_memory_methods}, we summarize twelve representative agent memory methods.} We categorize them according to four key dimensions: information extraction mechanism, memory management strategy, underlying storage structure, and retrieval method.
After a careful literature review, we make the following observations. 
\revise{First, there is a lack of a unified framework for systematically analyzing the implementation components of representative agent memory methods. }
Second, most previous studies report overall performance results, but rarely examine the roles and effects of individual components within these methods. Third, comprehensive and systematic comparisons among different methods—especially regarding their accuracy and efficiency—are still lacking.

\textbf{Our work.} We address these gaps by proposing a unified, modular framework and conducting an in-depth experimental study of representative agent memory methods. 
The framework decomposes memory mechanisms into four stages, including {\Large \ding{182}} {\textit{Information Extraction}},  {\Large \ding{183}} {\textit{Memory Management}}, {{{\Large \ding{184}}} {\it Memory Storage}, and {{{\Large \ding{185}}} {\it Information Retrieval}. Under this framework, we compare representative methods on two typical long-term conversational benchmarks, \textbf{LOCOMO}~\cite{maharana2024locomo} and \textbf{LONGMEMEVAL}~\cite{wu2025longmemeval}, \revise{as well as the agentic memory benchmark \textbf{MemoryArena}~\cite{he2026memoryarena}}. Beyond overall performance, we analyze practical robustness dimensions, including context scalability and positional sensitivity. 
Based on these analyses, we further design a new agent memory method that achieves the best overall performance and cost efficiency.

In summary, our principal contributions are listed as follows:
\begin{itemize}[topsep=1mm, partopsep=0pt, itemsep=0pt, leftmargin=10pt]
    \item \revise{We propose a unified framework that decomposes representative agent memory methods into four core memory modules,} enabling systematic comparisons of their differences.
    \item \revise{We conduct comprehensive experimental studies across conversational and agentic settings using LOCOMO, LONGMEMEVAL, and MemoryArena, together with analyses of efficiency}, context scalability, evidence position sensitivity, and LLM backbone dependence.
    \item \revise{Based on the above analyses, we propose a new agent memory method that achieves state-of-the-art performance among the evaluated representative baselines.} We further derive several key insights and highlight promising research directions for future work.
\end{itemize}


\textbf{Roadmap.} Section~\ref{sec:pre} introduces preliminaries. Section~\ref{sec:generic} presents the unified framework. Sections~\ref{sec:mem_extraction}--\ref{sec:mem_retrieval} characterize representative design choices within this framework. Section~\ref{sec:experiments} reports experimental results and analyses. Section~\ref{sec:lessons_opp} summarizes lessons and opportunities. Section~\ref{sec:related} reviews related work, and Section~\ref{sec:conclusions} concludes.

\section{Preliminaries}
\label{sec:pre}





In this section, we go through some important concepts and the typical workflow of existing memory methods in the LLM era. The relationship between RAG and memory is also discussed.

\subsection{LLM-related Concepts}
We introduce two fundamental LLM-related concepts below. 

{\bf LLM Prompting.}
Prompting~\cite{brown2020languagemodelsfewshotlearners, dong2024surveyincontextlearning,liu2021pretrainpromptpredictsystematic} specifies an LLM task by constructing an input context (prompt) that contains task instructions, the current input, and optionally demonstrations or auxiliary evidence. The model's output is then generated conditioned on this context. 

Prompting is particularly important in LLM-based systems because it provides a lightweight, training-free interface for task adaptation~\cite{liu2021pretrainpromptpredictsystematic}. Model behavior can be redirected by revising the input context without modifying model parameters. This property has enabled a wide range of practical applications in which task specifications and constraints are expressed directly in natural language~\cite{ouyang2022training,achiam2023gpt}.
In addition to final response generation, prompting is often used to drive intermediate subtasks in an LLM-centered pipeline~\cite{yao2023reactsynergizingreasoningacting,schick2024toolformer,asai2023self, khattab2023dspy}. Typical examples include extracting key information, consolidating intermediate results, and so on.

{\bf LLM-based Agents.}
An LLM-based agent utilizes an LLM as a core decision model for sequential action selection~\cite{yao2023reactsynergizingreasoningacting,shinn2023reflexion,li2024reviewprominentparadigmsllmbased}. In contrast to single-turn prompting, an agent operates in a closed loop: it receives observations, then performs reasoning or planning, executes an action (possibly via tools), obtains feedback from the environment, and further proceeds to the next step~\cite{yang2026graphbasedagentmemorytaxonomy}. The interaction history and available state are combined into textual context, based on which the agent predicts the next action. 

The action space of an LLM-based agent typically includes both natural-language responses and structured tool calls~\cite{packer2023memgpt, li2025memos} (e.g., information search, API invocation and memory read/write operations), which enable multi-step task completion beyond one-shot generation. Therefore, the LLM-based agent must retain and reuse information across conversation turns and sessions, which motivates effective memory mechanisms.




\begin{figure}[]
    \centering
    \setlength{\abovecaptionskip}{0cm}
    \setlength{\belowcaptionskip}{-0.3cm}
    \includegraphics[width=0.8\linewidth]{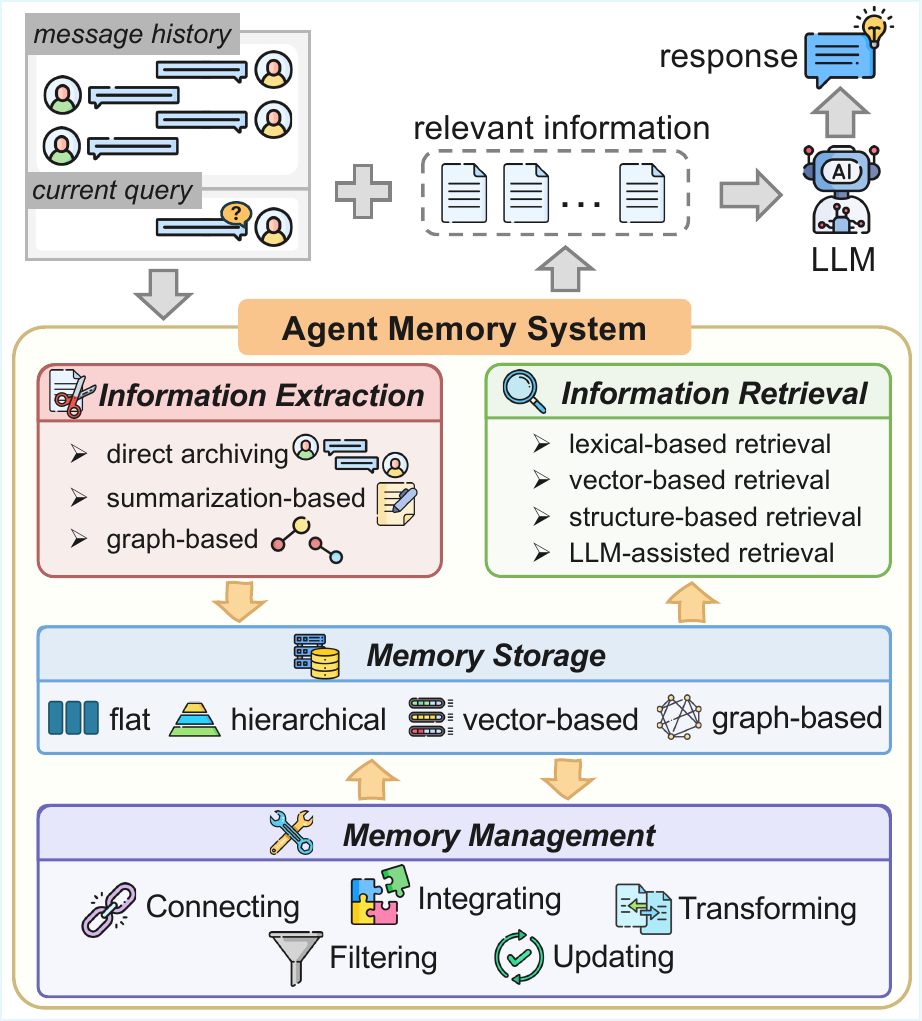}
    \caption{An overview of the unified framework for agent memory systems.}
    \label{fig:overview}
\end{figure}

\subsection{Agent Memory}


Memory is incorporated into LLM-based agents to compensate for the bounded context window~\cite{liu2024lost,maharana2024locomo, hsieh2024ruler}. Since the model conditions only on a limited number of tokens, information that comes from earlier turns and lies outside the current prompt can be easily lost, which degrades performance in long-horizon dialogue and multi-session tasks~\cite{ shaham-etal-2023-zeroscrolls, wu2025longmemeval, xu-etal-2022-beyond}. An explicit memory module~\cite{park2023generativeagents, packer2023memgpt, sumers2024cognitive, graves2016hybrid, bai-etal-2024-longbench} allows the system to persist interaction-derived information—such as user preferences, salient events, intermediate decisions, and task constraints—and to reintroduce it when relevant, thereby improving consistency and enabling reasoning that depends on long-term context~\cite{park2023generativeagents, graves2016hybrid}.

The typical workflow of memory-augmented systems begins by selectively extracting important information from ongoing interactions—such as facts, user preferences, or important events—and storing them as memory entries. These entries then undergo a series of management operations: consolidation~\cite{zhong2024memorybank, du2025rethinkingmemoryllmbased,10.5120/ijca2026926236}, which integrates similar memories to improve coherence and reduce redundancy; updating~\cite{hu2023chatdbaugmentingllmsdatabases, packer2023memgpt, liu2025dynamem}, which modifies stored content to maintain accuracy and reflect the latest knowledge; filtering~\cite{zhong2024memorybank,park2023generativeagents,li-etal-2023-compressing}, which removes outdated, redundant, or low-utility memories to preserve the efficiency and relevance of the system; and enhancement~\cite{hu-etal-2025-hiagent,sun-etal-2026-h}, which marks or surfaces important memories for easy identification and retrieval. This structured process ensures that the memory system remains organized, scalable, and aligned with ongoing user needs. When additional context is required for reasoning or generation, the memory system retrieves and supplies the most relevant information to the LLM, supporting both short-term adaptation and long-term continuity across evolving interactions. 


Memory and retrieval-augmented generation (RAG)~\cite{lewis2020retrieval, gutierrez2024hipporag, wang2024llmagentsurvey,gao2023retrieval} are related but distinct mechanisms. Memory primarily targets stateful, interaction-dependent information that evolves over time and is required for personalization and cross-session continuity~\cite{zhong2024memorybank, packer2023memgpt}. In contrast, RAG primarily targets external knowledge grounding, retrieving evidence from document collections or knowledge bases to supplement domain knowledge and reduce hallucinations~\cite{lewis2020retrieval, gao2023retrieval}. In practice, they are complementary: memory supplies user- and session-specific context, while RAG provides task-relevant factual evidence from external corpora.

\section{A UNIFIED FRAMEWORK}
\label{sec:generic}

\revise{In this section, we decompose existing agent memory systems into memory modules under a unified framework, as illustrated in Figure~\ref{fig:overview}.}
The framework comprises four core memory modules:
{\Large \ding{182}} \textit{Information extraction},
{\Large \ding{183}} \textit{Memory management},
{\Large \ding{184}} \textit{Memory storage}, and
{\Large \ding{185}} \textit{Information retrieval}.
\revise{Taken together, these modules capture how existing agent memory systems operate in practice.}

\revise{
As shown in Algorithm~\ref{alg:framework}, at interaction step $t$, let $\mathcal{M}_t$ denote the incoming messages, $q_t$ the current query, $\mathcal{D}_{t-1}$ the existing logical memory records, and $\mathcal{H}_{t-1}$ the materialized memory state. 
The four modules are formalized as
\[
\begin{aligned}
\mathcal{E}_t &= \texttt{Extract}_{\theta_E}(\mathcal{M}_t), &
\Delta_t &= \texttt{Manage}_{\theta_M}(\mathcal{H}_{t-1},\mathcal{E}_t),\\
\mathcal{H}_t &= \texttt{Materialize}_{\theta_S}(\mathcal{H}_{t-1},\Delta_t), &
\mathcal{X}_t &= \texttt{Retrieve}_{\theta_R}(q_t,\mathcal{H}_t).
\end{aligned}
\]
Specifically, \ding{182} \emph{information extraction} performs data ingestion and transformation, converting raw messages and their associated metadata, such as timestamps, into logical memory records $\mathcal{E}_t$. 
It filters out redundant details and transforms the relevant content into different types of knowledge (e.g., triples derived from text, informational summaries) suitable for downstream processing.
\ding{183} \emph{Memory management} performs incremental maintenance over the existing memory state $\mathcal{H}_{t-1}$.
Given the extracted records $\mathcal{E}_t$, it produces a change set $\Delta_t$ that may connect related records, integrate fragmented information, migrate records across memory levels, update existing content, or remove redundant and obsolete entries. 
\ding{184} \emph{Memory storage} materializes $\Delta_t$ into the updated persistent state
$\mathcal{H}_t=\langle\mathcal{D}_t,\mathcal{P}_t,\mathcal{I}_t\rangle$.
Here, $\mathcal{D}_t$ contains the logical memory records and their associated metadata, $\mathcal{P}_t$ specifies their flat or hierarchical organization, and $\mathcal{I}_t$ maintains vector- or graph-based representations together with their associated indexes and access structures.
\ding{185} \emph{Information retrieval} performs query processing over the materialized memory state $\mathcal{H}_t$. 
Given the current query $q_t$, it identifies candidate records, accesses them through the corresponding indexes or structures, and ranks their relevance to produce the retrieved context $\mathcal{X}_t$ for downstream reasoning or response generation.
}

\begin{algorithm}[t]
\revise{
  \caption{Data management workflow of the unified agent memory framework}
  \label{alg:framework}
  \small
  \SetKwInOut{Input}{Input}
  \SetKwInOut{Output}{Output}

  \Input{Incoming messages $M_t$, query $q_t$, persistent memory state
  $\mathcal{H}_{t-1}=\langle\mathcal{D}_{t-1},
  \mathcal{P}_{t-1},\mathcal{I}_{t-1}\rangle$, and module
  configurations $\Theta$}
  \Output{Updated memory state $\mathcal{H}_t$, retrieved context
  $\mathcal{X}_t$, and response $\mathcal{R}_t$}

  \tcp{\textcolor{teal}{{\Large \ding{182}} Information extraction}}
  $\mathcal{E}_t \gets
  \texttt{Extract}_{\theta_E}(M_t)$
  \tcp*[r]{construct logical memory records and metadata}

  \tcp{\textcolor{teal}{{\Large \ding{183}} Memory management}}
  $\Delta_t \gets
  \texttt{Manage}_{\theta_M}(\mathcal{H}_{t-1},\mathcal{E}_t)$
  \tcp*[r]{connect, integrate, transform, update, or filter records}

  \tcp{\textcolor{teal}{{\Large \ding{184}} Memory storage}}
  $\mathcal{H}_t \gets
  \texttt{Materialize}_{\theta_S}(\mathcal{H}_{t-1},\Delta_t)$
  \tcp*[r]{persist records and maintain organization and indexes}

  \tcp{\textcolor{teal}{{\Large \ding{185}} Information retrieval}}
  $\mathcal{X}_t \gets
  \texttt{Retrieve}_{\theta_R}(q_t,\mathcal{H}_t)$
  \tcp*[r]{retrieve and rank relevant memory records}

  $\mathcal{R}_t \gets
  \texttt{GenerateResponse}(q_t,\mathcal{X}_t)$\;

  \Return{$(\mathcal{H}_t,\mathcal{X}_t,\mathcal{R}_t)$}
}
\end{algorithm}


\section{Information extraction}
\label{sec:mem_extraction}

\revise{
This module serves to identify and extract information from $\mathcal{M}_t$ that is both useful and necessary for downstream memory processing.
}
As illustrated in Figure~\ref{fig:extraction}, existing agent systems adopt different information extraction methods, which can be broadly categorized as follows:

\begin{figure}[h]
    \centering
    \setlength{\abovecaptionskip}{0cm}
    \includegraphics[width=0.7\linewidth]{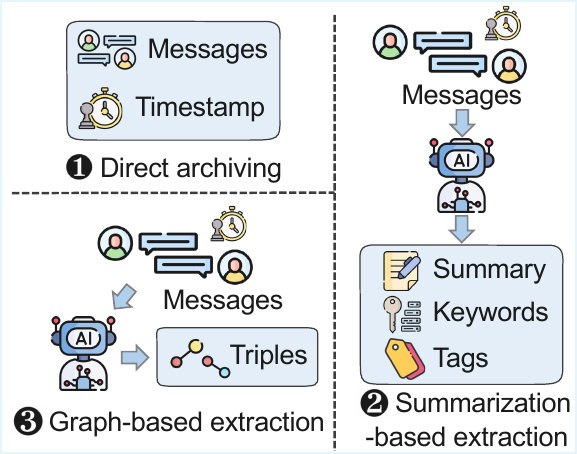}
    \caption{Methods of information extraction.}
    \label{fig:extraction}
\end{figure}

\begin{figure}[t]
\centering
\setlength{\abovecaptionskip}{0cm}
\begin{AIbox}{Prompt for Summarization-based Extraction.}
{
\begin{tcolorbox}[colback=gray!10, colframe=gray!10, boxrule=0pt, sharp corners, left=0pt, right=0pt]
Generate a structured analysis of the following message by extracting salient keywords, identifying core themes and contextual elements, and assigning relevant categorical tags.

Format the response as a JSON object: 

{\ttfamily
\{
    
    {\hangindent=2em \hangafter=0 "summary": "<one-sentence gist>", \par}

    \hspace*{2em}"keywords": [
    
        {\hangindent=4em \hangafter=0 <several keywords capturing key concepts> \par}

    \hspace*{2em}],
    
    \hspace*{2em}"tags": [
    
        {\hangindent=4em \hangafter=0 <several broad themes for classification> \par}

    \hspace*{2em}]
    
\}
}

Message for analysis:

\texttt{\{message\}}
\end{tcolorbox}
}
\end{AIbox}
\caption{A sample prompt for summarization-based extraction.}
\label{fig:pmt:summary_extract}
\end{figure}

\noindent {\Large \ding{182}} \textbf{Direct archiving.}
This method represents the most straightforward form of information extraction, where the agent system simply archives raw messages and timestamps without any processing.

\noindent {\Large \ding{183}} \textbf{Summarization-based extraction.}
This method employs LLMs to generate concise informational summaries from one or more dialogue turns. \revise{Memory methods such as {\tt A-MEM} and {\tt Mem0} extract keywords and contextual tags from $\mathcal{M}_t$,} or prompt the LLM to produce an abstracted summary of the raw text. Figure~\ref{fig:pmt:summary_extract} illustrates a representative prompt used for this extraction method.

\noindent {\Large \ding{184}} \textbf{Graph-based extraction.}
\revise{This method leverages LLMs to extract fine-grained entities and relations from $\mathcal{M}_t$,} forming subject–predicate–object triples for knowledge graph construction (e.g., {\tt Mem0$^g$}, {\tt Zep}). Additionally, temporal metadata such as creation or invalidation timestamps are recorded to support dynamic updates and temporal reasoning within the graph-based memory. A concrete example of the prompt designed for graph-based extraction is provided in Appendix~\ref{sec:app:prompt}.

\section{Memory Management}
\label{sec:mem_management}

\begin{figure*}[]
    \centering
    \setlength{\abovecaptionskip}{0cm}
    \setlength{\belowcaptionskip}{-0.3cm}
    \includegraphics[width=0.82\linewidth]{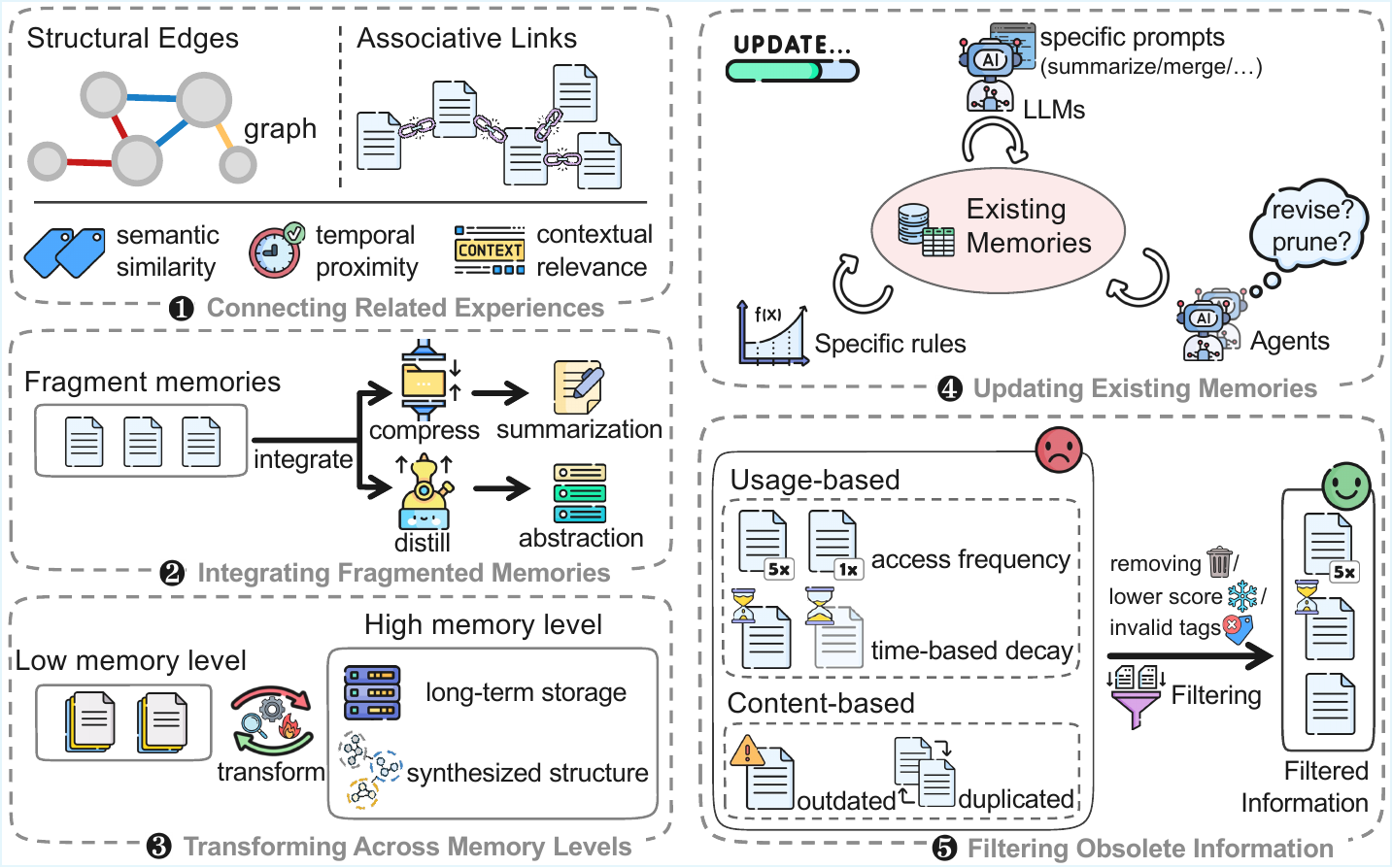}
    \caption{Workflow of the memory management process.}
    \label{fig:management}
\end{figure*}

\revise{
Memory management incrementally maintains and refines the existing logical memory records $\mathcal{D}_{t-1}$ using newly extracted records $\mathcal{E}_t$, producing a change set $\Delta_t$ that governs how memory evolves over time.
}
As illustrated in Figure~\ref{fig:management}, it mirrors the human memory lifecycle, encompassing five core operations: connecting related experiences, integrating fragmented information, transforming short-term into long-term memory, updating outdated content, and filtering obsolete knowledge. 
Table~\ref{tab:management_methods} summarizes how representative agent memory methods instantiate these operations through different implementation paradigms. 
Through this process, the system maintains a coherent, efficient, and adaptive memory state that supports continual learning and reasoning.

\noindent {\Large \ding{182}} \textbf{Connecting Related Experiences.}  
Humans naturally associate related events across time and context; agent systems emulate this through \textit{connection}.  
This mechanism establishes explicit connections between memory entries that share semantic similarity, temporal proximity, or contextual relevance, realized through either \textbf{structural edges} within a graph or \textbf{associative links} across discrete records.
For example, memory methods such as {\tt A-MEM} and {\tt MemoryOS} leverage associative links based on semantic similarity or continuity, enabling synchronous updates and alignment across connected memories. Separately, graph-based methods such as {\tt Zep} and {\tt Mem0$^g$} focus on connecting individual episode or entity nodes to support reasoning and retrieval across conceptually or temporally aligned memories.

\noindent {\Large \ding{183}} \textbf{Integrating Fragmented Memories.}  
Humans tend to summarize daily experiences, retaining only key events while discarding details.
Agent systems achieve similar \textit{integration} through \textbf{abstraction} or \textbf{summarization}. 
For example, {\tt MemoryBank} aggregates repetitive daily records into event summaries and refines a global user profile as experiences accumulate.  
This process reduces redundancy, distills essential information, and transforms scattered memories into concise high-level representations suitable for long-term storage.

\noindent {\Large \ding{184}} \textbf{Transforming Across Memory Levels.}  
Human memory gradually transfers important information from low-level to high-level storage, reinforcing what is repeatedly recalled.  
Agent systems adopt similar hierarchical migration mechanisms.  
For instance, {\tt MemoryOS} implements a two-stage migration strategy: short-term memories are first moved to mid-term storage following a First-In, First-Out (FIFO) policy, and mid-term memories are then promoted to long-term storage using a heat-based score that jointly considers access frequency and recency.  
Additionally, memory methods such as {\tt Zep} organize semantically related memories into communities, forming structured, interconnected long-term representations.  
This stage strengthens persistent knowledge while maintaining efficiency.

\noindent {\Large \ding{185}} \textbf{Updating Existing Memories.}  
Humans constantly revise their memories by integrating new experiences and correcting inconsistencies.  
Agent systems follow a similar principle through three main updating paradigms:
(1) \textbf{Rule-based updating}, where existing memories are updated according to predefined rules. For example, {\tt MemoryBank} adopts Ebbinghaus's Forgetting Curve theory to adjust memory strength over time. In {\tt MemoryOS}, new memories are integrated into existing structures based on semantic and keyword similarities;
(2) \textbf{LLM-based updating}, where large language models are prompted to summarize, merge, or resolve conflicts between entries. To give an example, {\tt MemTree} updates its memory by relying on the LLM to execute a specialized Aggregate Operation, where the prompt and the count of descendants guide the LLM to appropriately compress and generalize information before the new content is written back to the parent node. Taking another example, the update process in {\tt Zep} requires the LLM to execute resolution tasks by strictly following detailed semantic constraints and procedural guidelines given in the prompts.
(3) \textbf{Agent-based updating}, where agents autonomously decide which operations (e.g., revise, merge, prune) to apply, as in {\tt MemGPT} and {\tt MemOS}. To be more specific, agents are granted access to current context as well as historical or archival memory entries and learn to utilize specialized system tools for managing memories efficiently and flexibly.
These strategies ensure that memory remains accurate, consistent, and aligned with evolving knowledge.

\noindent {\Large \ding{186}} \textbf{Filtering Obsolete Information.}  
Finally, the memory system must remain compact and relevant by filtering outdated or redundant information, which can be achieved by either directly removing the memories, lowering their assigned weight score, or applying a status label (such as ``invalid''). This \textit{filtering} process parallels human forgetting, which selectively fades unused or irrelevant memories.  
(1) \textbf{Usage-based filtering}, as seen in {\tt MemoryOS} and {\tt MemoryBank}, relies on access frequency and time-based decay. Memories created a long time ago and rarely retrieved are filtered first.  
(2) \textbf{Content-based filtering} examines semantic similarity and leverages LLMs to detect and filter duplicated or outdated knowledge, such as in {\tt Mem0} and {\tt Mem0$^g$}, reducing noise and improving retrieval precision.  
Together, these mechanisms sustain an efficient, lightweight memory that supports ongoing adaptation and learning.

\begin{table*}[]
  \centering
  \caption{\revise{Comparison of implementation paradigms for memory management operations; ``N/A'' means that this operation is not explicitly implemented.}}
  \small
  \label{tab:management_methods}
  \revise{
  \begin{tabular}{l|c|c|c|c|c}
     \toprule
     \rowcolor{gray!20}
      \textbf{Method} & \textbf{Connecting} & \textbf{Integrating} & \textbf{Transforming} & \textbf{Updating}& \textbf{Filtering} \\
     \midrule
      {\tt A-MEM} & Associative Links & N/A & N/A & LLM-based & N/A \\
      {\tt MemoryBank} & N/A & Summarization & N/A & Rule-based & Usage-based \\
      {\tt MemGPT} & N/A & Summarization & Stage-wise Transfer & Agent-based & N/A \\
      {\tt Mem0} & N/A & Summarization & N/A & Agent-based & Content-based \\
      {\tt Mem0$^g$} & Structural Edges & N/A & N/A & LLM-based & Content-based \\
      {\tt MemoChat} & N/A & Abstraction & N/A & N/A & N/A \\
      {\tt Zep} & Structural Edges & N/A & Community Formation & LLM-based & N/A \\
      {\tt MemTree} & Structural Edges & Summarization & N/A & LLM-based & N/A \\
      {\tt MemoryOS} & Associative Links & Abstraction + Summarization & Stage-wise Transfer & Rule-based & Usage-based \\
      {\tt MemOS} & Structural Edges & Abstraction + Summarization & N/A & Agent-based & N/A \\
      {\tt MemGAS} & Structural Edges & N/A & N/A & N/A & N/A \\
      {\tt LightMem} & N/A & Summarization & N/A & Agent-based & Content-based \\
     \bottomrule
  \end{tabular}
  }
\end{table*}

\section{Memory Storage}
\label{sec:mem_storage}

\revise{
Memory storage materializes the change set $\Delta_t$ into the updated persistent memory state $\mathcal{H}_t$, governing how logical memory records $\mathcal{D}_t$ are organized and represented along two primary dimensions: organization-centric and representation-centric.
}
The former determines the architectural depth of the storage system, encompassing flat storage and hierarchical storage, and the latter characterizes the employed technological paradigm, primarily comprising vector-based storage and graph-based storage.

\noindent {\Large \ding{182}} \textbf{Flat storage.}
This approach represents a unified, single-tier storage that aggregates all information within a homogeneous space, such as a FIFO queue or a JSON file, defined relative to hierarchical storage within the organization-centric dimension.

\noindent {\Large \ding{183}} \textbf{Hierarchical storage.}
This approach partitions memory into specialized, multi-tiered architectures, allowing individual storage components to fulfill distinct functional roles and operate at different levels of granularity. For example, {\tt MemoryOS} organizes memory into a three-tier hierarchical structure: short-term memory for timely conversations, mid-term memory for topic summaries, and long-term memory for user preferences. 
By applying different, synergistic management and retrieval strategies to respective storage components, hierarchical storage effectively optimizes the trade-off between computational overhead and knowledge persistence.

\noindent {\Large \ding{184}} \textbf{Vector-based storage.}
This approach encodes textual memory into high-dimensional embeddings, subsequently indexed in dedicated vector libraries or databases, such as FAISS~\cite{douze2025faisslibrary} and Qdrant, to enable the agent to perform efficient semantic similarity search. Vector-based storage can function as a standalone repository or serve as a foundational building block frequently integrated into more complex storage architectures.

\noindent {\Large \ding{185}} \textbf{Graph-based storage.}
This approach utilizes diverse graph topologies, such as trees, knowledge graphs, and temporal graphs, to preserve the rich structural information inherent in memory. For instance, {\tt MemTree} organizes memory into a hierarchical tree where each node encapsulates aggregated textual content, providing varying levels of abstraction along the tree's depth; {\tt Zep} employs a layered temporal knowledge graph that concurrently organizes memory by representing raw messages as nodes, extracting subject–predicate–object triples, and clustering entities into communities. These graph-based storage methods capture intricate relationships and multi-hop associations that lie beyond the reach of simple vector similarity metrics.
\section{Information retrieval}
\label{sec:mem_retrieval}

\revise{
This module processes the incoming query $q_t$ over the persistent memory state $\mathcal{H}_t$, identifying and ranking relevant memory records to produce the retrieved context $\mathcal{X}_t$ for downstream reasoning or context-aware response generation.
}
Existing information retrieval strategies can be broadly categorized into four paradigms based on the fundamental mechanisms they employ:

\noindent {\Large \ding{182}} \textbf{Lexical-Based Retrieval.}
This paradigm relies on the overlap of surface-level tokens or terms, typically implemented through some representative techniques such as set-based matching via the Jaccard similarity coefficient or scoring models like BM25~\cite{robertson2009prf}. Lexical-based retrieval provides a strong baseline for exact term matching, which can be particularly effective for retrieving names, specific entities, or phrases where precise wording is critical.

\noindent {\Large \ding{183}} \textbf{Vector-Based Retrieval.}
This paradigm leverages semantic similarity in a continuous vector space to address the vocabulary mismatch problem inherent in exact keyword matching. By encoding both the query and memories into high-dimensional vectors via embedding models, vector-based retrieval is formulated as a top-$k$ search for the most relevant entries using distance metrics like cosine similarity. This approach excels at capturing latent semantic nuances, ensuring that relevance is determined by semantic content rather than surface-level lexical form. To maintain efficiency within the massive scale of memory storage, Approximate Nearest Neighbor (ANN) search algorithms, such as HNSW~\cite{malkov2018efficient} or PQ~\cite{jegou2011pq} (Product Quantization), are frequently employed.

\noindent {\Large \ding{184}} \textbf{Structure-Based Retrieval.}
This paradigm exploits the explicit relational connections between memory entities, often operating on graph-based or hierarchical storage, performing graph traversal, neighborhood expansion, or multi-hop reasoning to retrieve interconnected clusters of information instead of simple query-to-item matching. For example, {\tt Mem0$^g$} explores the relationships starting from nodes identified through similarity search to construct a comprehensive subgraph that captures relevant and multi-faceted information. Similarly, {\tt Zep} utilizes a BFS-based graph traversal algorithm to enhance initial search results by identifying additional nodes and edges.

\noindent {\Large \ding{185}} \textbf{LLM-Assisted Retrieval.}
This paradigm integrates LLMs as an active reasoning component to guide or refine the retrieval process. In addition to directly deciding which specific information should be retrieved, LLMs can also be utilized to transform ambiguous user prompts into precise search queries or to identify key entities within a query to facilitate more targeted retrieval. By leveraging the reasoning capabilities of LLMs, this paradigm excels at uncovering latent semantic dependencies, thereby ensuring closer alignment between queries and retrieved knowledge.

\section{Experiments}
\label{sec:experiments}

We now present the experimental results. We discuss the setup in Section~\ref{sec:exp:setup}, and then report the evaluation results across several experiments in Section~\ref{sec:exp:eval}.

\definecolor{sota}{RGB}{191, 0, 64}
\definecolor{second}{RGB}{191, 129, 64}
\newcommand{\fst}[1]{\textbf{\textcolor{sota}{#1}}}
\newcommand{\snd}[1]{\textbf{\textcolor{second}{#1}}}

\subsection{Setup}
\label{sec:exp:setup}

\begin{table*}[t]
\centering
\caption{\revise{Comparison of methods on LONGMEMEVAL. The best and second-best results are marked in bold and underlined, respectively.}}
\small
\begin{threeparttable}
\revise{
\begin{tabular}{l|r|r|r|r|r|r|r|r|r|r|r|r|r|r}
\toprule
\multicolumn{1}{l}{\multirow{3}[2]{*}{Method}} & \multicolumn{6}{c}{Information Extraction} & \multicolumn{2}{c}{\multirow{2}[1]{*}{Multi-Session}} & \multicolumn{2}{c}{\multirow{2}[1]{*}{Temporal}} & \multicolumn{2}{c}{\multirow{2}[1]{*}{Knowledge Updates}} & \multicolumn{2}{c}{\multirow{2}[1]{*}{Overall}}\\ [-2pt] \cmidrule(lr){2-7}
 \multicolumn{1}{l}{} & \multicolumn{2}{c}{user}  & \multicolumn{2}{c}{assistant} & \multicolumn{2}{c}{preference} & \multicolumn{2}{c}{} & \multicolumn{2}{c}{} & \multicolumn{2}{c}{} \\ [-2pt] \cmidrule(lr){2-3} \cmidrule(lr){4-5} \cmidrule(lr){6-7} \cmidrule(lr){8-9} \cmidrule(lr){10-11} \cmidrule(lr){12-13} \cmidrule(lr){14-15}
 \multicolumn{1}{l}{} & \multicolumn{1}{r}{F1} &  \multicolumn{1}{r}{BLEU-1} & \multicolumn{1}{r}{F1} &  \multicolumn{1}{r}{BLEU-1}  & \multicolumn{1}{r}{F1} &  \multicolumn{1}{r}{BLEU-1} & \multicolumn{1}{r}{F1} &  \multicolumn{1}{r}{BLEU-1} & \multicolumn{1}{r}{F1} &  \multicolumn{1}{r}{BLEU-1} & \multicolumn{1}{r}{F1} &  \multicolumn{1}{r}{BLEU-1} & \multicolumn{1}{r}{F1} &  \multicolumn{1}{r}{BLEU-1}\\ \midrule
 \multicolumn{15}{c}{Qwen3.5-9B}\\ \midrule
{\tt A-MEM}&40.15 &35.99&43.51 &36.02&11.72 &0.91&17.08 &14.58&22.91&16.47&31.91 &29.05&26.81 &21.92\\
{\tt MemoryBank}&56.17 &46.14&66.64 &55.02&\textbf{20.67} &\textbf{9.54}&17.70 &10.73&22.41 &14.75&37.20 &27.84&33.04 &24.32\\
{\tt MemGPT} &55.54 & 49.79 & 54.67 & 46.00 & 13.63 & 3.82 & 19.19 & 17.19 & 22.30 & 15.86 & 29.54 & 27.07 & 30.36 & 25.37\\ 
{\tt Mem0}&56.46 &47.35&47.78 &37.97&9.03 &0.04&25.52 &20.73&29.65&18.42&36.70 &27.54&34.20 &25.59\\
{\tt Mem0\(^g\)} &55.56 & 48.44 & 22.93 & 18.25 & 11.85 & 0.47 & \underline{30.71} & \textbf{28.41} & 35.45 & \underline{27.13} & 40.86 & 33.42 & 35.03 & 28.84\\ 
{\tt MemoChat} &12.37 & 6.01 & 31.00 & 23.37 & 10.69 & 0.91 & 21.85 & 17.94 & 31.36 & 21.03 & 11.25 & 8.41 & 21.75 & 15.19\\ 
{\tt Zep} &--- &--- &--- &--- &--- &--- &--- &--- &--- &--- &--- &--- &--- &---\\ 
{\tt MemTree} &\underline{68.18} & \underline{63.67} & 58.82 & 45.94 & 11.46 & 1.57 & 26.05 & 23.89 & \underline{36.61} & 26.07 & \underline{49.69} & \textbf{46.49} & \textbf{41.24} & \underline{34.69}\\ 
{\tt MemoryOS} &65.23 & 59.56 & \underline{66.67} & \underline{59.82} & 12.87 & 1.66 & 27.29 & 24.16 & 36.38 & \textbf{29.76} & 43.06 & 39.31 & \underline{41.02} & \textbf{35.61}\\
{\tt MemOS} &67.59 & 60.09 & 58.84 & 50.76 & \underline{14.28} & 0.55 & 26.81 & 23.95 & 25.91 & 18.34 & 40.74 & 33.81 & 37.29 & 30.66\\
{\tt MemGAS} &61.98 & 56.57 & \textbf{72.55} & \textbf{61.85} & 11.73 & 2.95 & 16.38 & 14.78 & 21.10 & 16.73 & 42.23 & 36.75 & 34.06 & 29.14\\
{\tt LightMem} &\textbf{70.12} & \textbf{65.05} & 18.33 & 13.97 & 13.13 & \underline{4.69} & \textbf{31.67} & \underline{28.07} & \textbf{39.33} & 27.05 & \textbf{49.76} & \underline{44.05} & 39.31 & 32.49\\     \midrule
\multicolumn{15}{c}{Qwen3.5-27B}\\ \midrule
{\tt A-MEM}&57.37 & 49.83 & 51.76 & 42.14 & 11.87 & 0.20 & 22.33 & 20.05 & 28.31 & 23.01 & 43.84 & 30.90 & 34.85 & 27.98\\ 
{\tt MemoryBank}&54.91 & 51.38 & 60.71 & 53.36 & \underline{13.62} & \textbf{4.59} & 34.71 & 29.79 & 29.92 & 26.00 & 39.52 & 31.35 & 38.66 & 33.18\\ 
{\tt MemGPT} &57.03 & 52.48 & 65.79 & 50.46 & 6.11 & 0.95 & 22.67 & 19.83 & 30.24 & 21.39 & 32.09 & 25.34 & 34.80 & 27.97\\ 
{\tt Mem0}&70.20 & 62.93 & 50.05 & 41.67 & \textbf{13.95} & 0.99 & 33.03 & 31.60 & 36.27 & 29.16 & 48.23 & 44.17 & 42.23 & 36.59\\ 
{\tt Mem0\(^g\)}&65.67 & 57.90 & 41.86 & 37.84 & 12.17 & 1.14 & 38.45 & 31.44 & 39.71 & 34.41 & 51.03 & 48.42 & 43.36 & 37.48\\ 
{\tt MemoChat} &19.41 & 16.42 & 27.13 & 21.37 & 8.33 & 0.36 & 25.26 & 22.43 & 32.40 & 22.28 & 22.96 & 20.29 & 25.17 & 19.77\\ 
{\tt Zep} & ---& ---& ---& ---& ---& ---& ---& ---& ---& ---& ---& ---& ---& ---\\ 
{\tt MemTree} &70.29 & 62.22 & 72.22 & 61.83 & 11.41 & 1.03 & \textbf{44.75} & \textbf{39.86} & \underline{40.68} & \underline{29.90} & 53.06 & 48.96 & \underline{49.62} & \underline{41.89}\\ 
{\tt MemoryOS} &\textbf{76.14} & \textbf{71.79} & \textbf{78.81} & \textbf{66.96} & 12.72 & 1.87 & \underline{44.44} & \underline{38.32} & 39.06 & 29.33 & \textbf{57.94} & \textbf{53.92} & \textbf{51.50} & \textbf{44.07}\\
{\tt MemOS} &72.14 & 66.07 & 56.48 & 45.35 & 13.38 & 0.55 & 35.55 & 30.49 & 37.14 & 22.72 & 53.80 & 44.27 & 44.96 & 35.42\\
{\tt MemGAS} &67.98 & 61.79 & \underline{73.52} & \underline{62.16} & 12.74 & 4.21 & 17.26 & 16.51 & 22.70 & 18.69 & 46.57 & 41.29 & 36.41 & 31.67\\
{\tt LightMem} &\underline{72.52} & \underline{66.25} & 21.94 & 16.86 & 13.14 & \underline{4.43} & 37.10 & 32.14 & \textbf{44.45} & \textbf{30.26} & \underline{57.06} & \underline{50.18} & 43.99 & 35.86\\ 
\bottomrule
\end{tabular}
\begin{tablenotes}[flushleft]
\small
\item[$\dagger$] The unavailable results for \texttt{Zep} are explained in Appendix~\ref{sec:app:zep}.
\end{tablenotes}
}
\end{threeparttable}
\label{tab:lme}
\end{table*}

\noindent \begin{tikzpicture}
\filldraw (0,0) -- (-0.15,0.08) -- (-0.15,-0.08) -- cycle ; 
\end{tikzpicture} \textbf{\underline{Workflow of our evaluation.}} 

We conduct a systematic experimental study of agent memory mechanisms along three dimensions:
\revise{(1) we collect and reimplement 12 representative methods within the unified framework described in Section~\ref{sec:generic};}
\revise{(2) we perform a comprehensive evaluation on two widely-used long-term conversational memory benchmarks and an agentic memory benchmark using multiple complementary metrics;}
(3) we carry out multi-dimensional analyses to assess architectural trade-offs and robustness, \revise{covering token cost and retrieval latency efficiency}, ground-truth position sensitivity, context scalability, and LLM backbone dependence.

\noindent \begin{tikzpicture}
\filldraw (0,0) -- (-0.15,0.08) -- (-0.15,-0.08) -- cycle ; 
\end{tikzpicture} \textbf{\underline{Benchmark Datasets.}} 
\revise{We employ three benchmark datasets to evaluate the performance of each memory mechanism. To assess long-term conversational memory capabilities, we use LOCOMO~\cite{maharana2024locomo} and LONGMEMEVAL~\cite{wu2025longmemeval}, which represent two distinct interaction scenarios: human--human dialogues and user--AI interactions, respectively. We further use MemoryArena~\cite{he2026memoryarena} to evaluate memory mechanisms in agent--environment interactions.}
Further dataset details are provided in Appendix~\ref{sec:app:exp:detail}. Due to the configurable structure of LONGMEMEVAL, we construct specific variants to evaluate \textbf{context scalability} and \textbf{position sensitivity}. The detailed variant construction methodology is outlined in Appendix~\ref{sec:app:exp:construct}.

\revise{
\noindent \begin{tikzpicture}
\filldraw (0,0) -- (-0.15,0.08) -- (-0.15,-0.08) -- cycle ; 
\end{tikzpicture} \textbf{\underline{Method Selection.}} 
We focus on text-centric agent memory methods and select representative approaches based on their relevance, recency, reproducibility, and architectural diversity. As summarized in Table~\ref{tab:image_memory_methods} and Table~\ref{tab:management_methods}, the selected methods cover the major design choices across the four memory modules. Further method selection details are provided in Appendix~\ref{sec:app:method_selection}.
}


\revise{
\noindent \begin{tikzpicture} 
\filldraw (0,0) -- (-0.15,0.08) -- (-0.15,-0.08) -- cycle ; 
\end{tikzpicture} \textbf{\underline{Evaluation Metrics.}}
For the conversational benchmarks, LOCOMO and LONGMEMEVAL, following their evaluation protocols and prior studies on long-horizon conversational memory, we adopt three complementary metrics.
\textbf{F1} measures token-level overlap by balancing precision and recall.
\textbf{BLEU-1} captures unigram-level modified precision with a brevity penalty, reflecting lexical fidelity to the reference answer.
\textbf{Accuracy} is assessed by an LLM judge~(GPT-5.4-mini) that determines whether the generated answer is semantically consistent with the reference answer.
The specific prompt for LLM-as-a-judge is detailed in Appendix~\ref{sec:app:prompt}.

For the agentic benchmark MemoryArena, we follow its official evaluation protocol and report \textbf{Task Success Rate (SR)}, which measures the proportion of fully completed tasks, and \textbf{Task Progress Score (PS)}, which measures partial task completion based on the proportion of satisfied task criteria. For \textit{Progressive Web Search}, we additionally report \textbf{Accuracy}, measured by the correctness of the final search query in each task, and \textbf{\#Search}, measured by the number of search-tool calls used to answer the final integrated query.
}

\begin{table*}[t]
\centering
\caption{\revise{Comparison of methods on LOCOMO. The best and second-best results are marked in bold and underlined, respectively.}}
\small
\revise{
\begin{tabular}{l|r|r|r|r|r|r|r|r|r|r|r|r|r|r|r}
\toprule
\multirow{2}{*}{Method} & \multicolumn{3}{c|}{Single-Hop}  & \multicolumn{3}{c|}{Multi-Hop} & \multicolumn{3}{c|}{Temporal}  & \multicolumn{3}{c|}{Open-Domain} & \multicolumn{3}{c}{Overall}\\ \cline{2-16}
 & F1 &  BLEU-1 & Acc & F1  & BLEU-1 & Acc & F1 & BLEU-1 & Acc & F1 & BLEU-1 & Acc & F1 & BLEU-1 & Acc \\ \midrule

\multicolumn{16}{c}{Qwen3.5-9B}\\ \midrule

{\tt A-MEM}
& 34.38 & 30.22 & 53.75 & 23.26 & 15.18 & 52.13 & 18.30 & 13.12 & 52.02
& 12.38 & 8.78 & 39.58 & 27.62 & 22.57 & 52.21 \\

{\tt MemoryBank}
& 45.05 & 39.09 & 71.70 & \underline{36.93} & \underline{28.77} & 78.01
& 27.29 & 22.20 & 54.52 & 13.83 & 10.05 & 38.54 & 37.92 & 31.87 & 67.21 \\

{\tt MemGPT}
& 38.97 & 33.32 & 54.34 & 25.11 & 17.45 & 50.00 & 16.57 & 13.47 & 41.74
& 14.99 & 10.99 & 34.38 & 30.27 & 24.88 & 49.68 \\

{\tt Mem0}
& 34.66 & 29.75 & 63.26 & 29.69 & 21.36 & 68.79
& \underline{42.68} & \underline{36.97} & 61.68
& 20.66 & 14.70 & 45.83 & 34.55 & 28.78 & 62.86 \\

{\tt Mem0\(^g\)}
& 26.41 & 21.76 & 58.50 & 21.07 & 14.59 & 57.09 & 39.35 & 33.33 & 62.93
& 15.76 & 10.50 & 48.96 & 27.47 & 22.15 & 58.57 \\

{\tt MemoChat}
& 6.64 & 5.12 & 9.63 & 8.07 & 5.16 & 18.09 & 4.94 & 4.10 & 4.67
& 9.75 & 6.08 & 26.04 & 6.74 & 4.98 & 11.17 \\

{\tt Zep}
& \underline{48.72} & \underline{42.98} & \textbf{81.23}
& 36.29 & 28.13 & \textbf{84.40}
& 40.90 & 35.96 & \underline{71.19}
& 17.84 & 13.25 & 51.04
& \textbf{42.89} & \textbf{36.94} & \textbf{77.84} \\

{\tt MemTree}
& 43.42 & 36.24 & 69.23 & 34.21 & 22.57 & 71.92 & 33.08 & 27.60 & 60.97
& \textbf{26.54} & \textbf{21.28} & \textbf{58.30}
& 38.53 & 31.01 & 72.85 \\

{\tt MemoryOS}
& 43.58 & 36.61 & 67.54 & 35.64 & 25.26 & 73.40 & 28.05 & 21.36 & 50.47
& 18.98 & 13.61 & \underline{56.25} & 37.36 & 29.92 & 64.35 \\

{\tt MemOS}
& 45.38 & 40.16 & 74.53
& \textbf{38.41} & \textbf{30.60} & \underline{80.50}
& \textbf{43.45} & \textbf{39.24} & \textbf{72.26}
& \underline{21.49} & 14.97 & 52.08
& \underline{42.21} & \underline{36.65} & \underline{73.75} \\

{\tt MemGAS}
& \textbf{51.55} & \textbf{45.26} & \underline{78.00}
& 36.28 & 28.43 & 76.60 & 15.33 & 11.42 & 36.76
& 19.47 & \underline{15.23} & 46.88 & 39.21 & 33.25 & 67.21 \\

{\tt LightMem}
& 37.93 & 31.87 & 68.37 & 25.00 & 17.24 & 64.54 & 40.01 & 33.09 & 68.22
& 17.22 & 11.48 & 52.08 & 34.70 & 28.18 & 66.62 \\

\midrule
\multicolumn{16}{c}{Qwen3.5-27B}\\ \midrule

{\tt A-MEM}
& 34.76 & 30.28 & 53.63 & 26.06 & 17.32 & 58.87 & 14.88 & 10.48 & 44.86
& 12.39 & 9.56 & 34.38 & 27.63 & 22.49 & 51.56 \\

{\tt MemoryBank}
& 46.56 & 40.24 & 75.51 & 40.38 & \underline{32.10} & 78.72
& 34.38 & 28.49 & 61.37 & 18.69 & 14.59 & 43.75 & 41.15 & 34.70 & 71.17 \\

{\tt MemGPT}
& 50.98 & 44.54 & 72.53 & 32.08 & 22.89 & 62.41 & 20.85 & 16.95 & 49.53
& 18.29 & 14.53 & 40.62 & 39.20 & 32.95 & 63.90 \\

{\tt Mem0}
& 41.11 & 35.38 & 71.94 & 36.96 & 28.13 & 74.47
& \underline{52.97} & \underline{46.33} & 74.45
& 22.73 & 17.25 & 50.00 & 41.68 & 35.20 & 71.56 \\

{\tt Mem0\(^g\)}
& 38.48 & 33.16 & 68.85 & 39.30 & 29.71 & 78.37 & 52.69 & 46.16 & 74.14
& \underline{26.58} & 19.76 & 57.29 & 40.85 & 34.40 & 70.97 \\

{\tt MemoChat}
& 6.93 & 5.46 & 8.80 & 6.52 & 3.66 & 14.18 & 7.08 & 6.05 & 4.98
& 10.98 & 7.22 & 22.92 & 7.14 & 5.36 & 9.87 \\

{\tt Zep}
& \textbf{53.15} & \textbf{46.29} & \textbf{84.37}
& 37.92 & 29.41 & \textbf{86.17}
& 45.19 & 39.66 & 74.45
& 19.68 & 15.14 & 55.21
& \underline{46.62} & \underline{39.88} & \textbf{80.81} \\

{\tt MemTree}
& 44.30 & 37.21 & 71.38 & 37.79 & 28.90 & 76.14 & 34.56 & 28.32 & 67.56
& 26.18 & \underline{19.97} & 57.91 & 39.95 & 32.76 & 75.64 \\

{\tt MemoryOS}
& 50.71 & 44.27 & 76.34 & \underline{40.39} & 30.09 & 79.43
& 37.22 & 29.27 & 62.31 & 24.20 & 18.16 & \textbf{61.46}
& 44.36 & 36.92 & 73.05 \\

{\tt MemOS}
& \underline{52.04} & 44.92 & \underline{82.33}
& 38.48 & 29.43 & 78.01
& \textbf{55.98} & \textbf{48.64} & \textbf{81.36}
& \textbf{27.97} & \textbf{20.72} & \underline{60.42}
& \textbf{48.88} & \textbf{41.35} & \underline{79.97} \\

{\tt MemGAS}
& 51.95 & \underline{45.73} & 76.10
& \textbf{40.97} & \textbf{32.33} & \underline{82.62}
& 25.52 & 21.64 & 39.88
& 21.18 & 15.93 & 48.96 & 42.51 & 36.40 & 68.05 \\

{\tt LightMem}
& 40.69 & 34.23 & 72.06 & 27.62 & 19.56 & 67.02
& 52.82 & 44.32 & \underline{80.06}
& 22.22 & 16.73 & 56.25 & 39.67 & 32.55 & 71.82 \\
\bottomrule
\end{tabular}
}
\label{tab:loco}
\end{table*}

\noindent  \begin{tikzpicture}
\filldraw (0,0) -- (-0.15,0.08) -- (-0.15,-0.08) -- cycle ; 
\end{tikzpicture} \textbf{\underline{Implementation.}}
We implement all methods in Python under the proposed unified framework, ensuring faithful and consistent reimplementation based on original papers and publicly available code. All experiments are conducted on 8 NVIDIA A100 (80\,GB) GPUs. If a method cannot finish in two days, we mark its result as ``---'' in the tables.
Additionally, since F1 and BLEU-1 metrics are highly sensitive to answer verbosity, we apply a uniform simplification step to all generated answers. \revise{The specific prompt and discussion for this process is detailed in Appendix~\ref{sec:app:prompt} and Appendix~\ref{sec:app:metric}, respectively.}

\noindent  \begin{tikzpicture}
\filldraw (0,0) -- (-0.15,0.08) -- (-0.15,-0.08) -- cycle ; 
\end{tikzpicture} \textbf{\underline{Hyperparameter Settings.}}
\revise{Unless otherwise stated, we use Qwen3.5-9B as the default LLM backbone, as it represents a capable and accessible open-source model.} The maximum context length is set to 20{,}000 tokens, and we employ greedy decoding to ensure deterministic outputs. For all methods that involve top-$k$ retrieval, following the previous work~\cite{kang2025memory,rezazadeh2024isolated,wujiang2025amem} \revise{and based on the sensitivity analysis in Section~8.2 (Exp.7), we set $k=10$ to balance retrieval effectiveness and context length.}
We adopt all-MiniLM-L6-v2, a representative and widely used sentence-transformer model, as the unified embedding model across all methods. All remaining method-specific hyperparameters follow the original settings reported in their respective papers and codebases.

\subsection{Evaluation}
\label{sec:exp:eval}

\begin{tikzpicture}
\filldraw (0,0) -- (-0.15,0.08) -- (-0.15,-0.08) -- cycle ; 
\end{tikzpicture} \textbf{Exp.1. Overall performance.}
We first report the performance of all agent memory methods on both benchmarks across two model scales \revise{(Qwen3.5-9B and Qwen3.5-27B)}. Results on LONGMEMEVAL and LOCOMO are shown in Table~\ref{tab:lme} and Table~\ref{tab:loco}, respectively. Based on these results, we draw the following observations:

(1) Tree-based memory methods (e.g., {\tt MemTree} and {\tt MemOS}) generally achieve strong performance by organizing memory in a multi-layered, multi-granularity fashion. \revise{Specifically, {\tt MemTree} achieves strong performance on LONGMEMEVAL, while {\tt MemOS} demonstrates competitive results on LOCOMO.} Tree structures provide high-level conceptual summaries at upper layers while preserving fine-grained details at leaf nodes. A similar advantage can be realized through well-designed hierarchical architectures that facilitate efficient information flow and transformation across different levels of abstraction, as evidenced by the highly competitive results of {\tt MemoryOS} and {\tt Zep}.

(2) Preserving information completeness is crucial for effective memory persistence---specifically, retaining raw messages during the information extraction phase and incorporating original conversations during final response generation. For example, methods that exclusively extract graph-based triples may suffer from information loss compared to those that preserve raw dialogue fragments, which may explain why {\tt Mem0} outperforms {\tt Mem0$^g$} in many cases.


\revise{
(3) Multi-session and temporal reasoning tasks remain highly sensitive to the reasoning capability of the backbone LLM. When scaling the backbone model from 9B to 27B, substantial performance improvements can be observed across different memory architectures. Specifically, on the Multi-Session category of LONGMEMEVAL, both {\tt MemTree} and {\tt MemoryOS} achieve more than $1.6\times$ performance gains, while on the Temporal category of LOCOMO, {\tt Mem0} and {\tt MemOS} improve their F1 scores by over 10 points. These results indicate that stronger LLM reasoning capabilities can significantly benefit memory-intensive tasks, while also motivating the development of memory architectures that provide explicit support for complex reasoning over persistent information.
}

\begin{figure}[t]
    \centering
    \setlength{\abovecaptionskip}{0cm}
    \setlength{\belowcaptionskip}{0cm}

    \begin{subfigure}{0.49\linewidth}
        \centering
        \includegraphics[width=\linewidth]{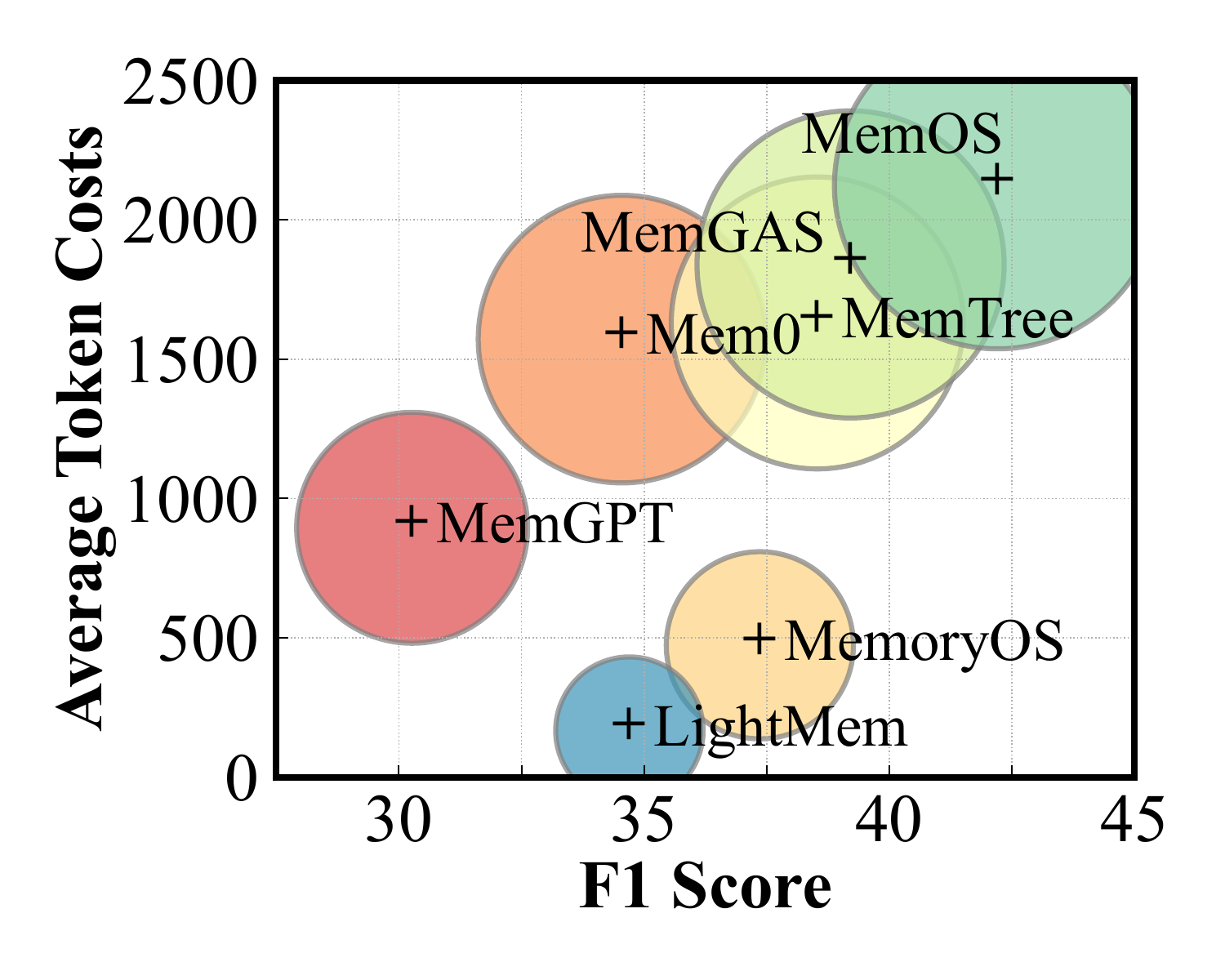}
        \caption{Trade-off}
        \label{fig:tradeoff}
    \end{subfigure}
    \hfill
    \begin{subfigure}{0.49\linewidth}
        \centering
        \includegraphics[width=\linewidth]{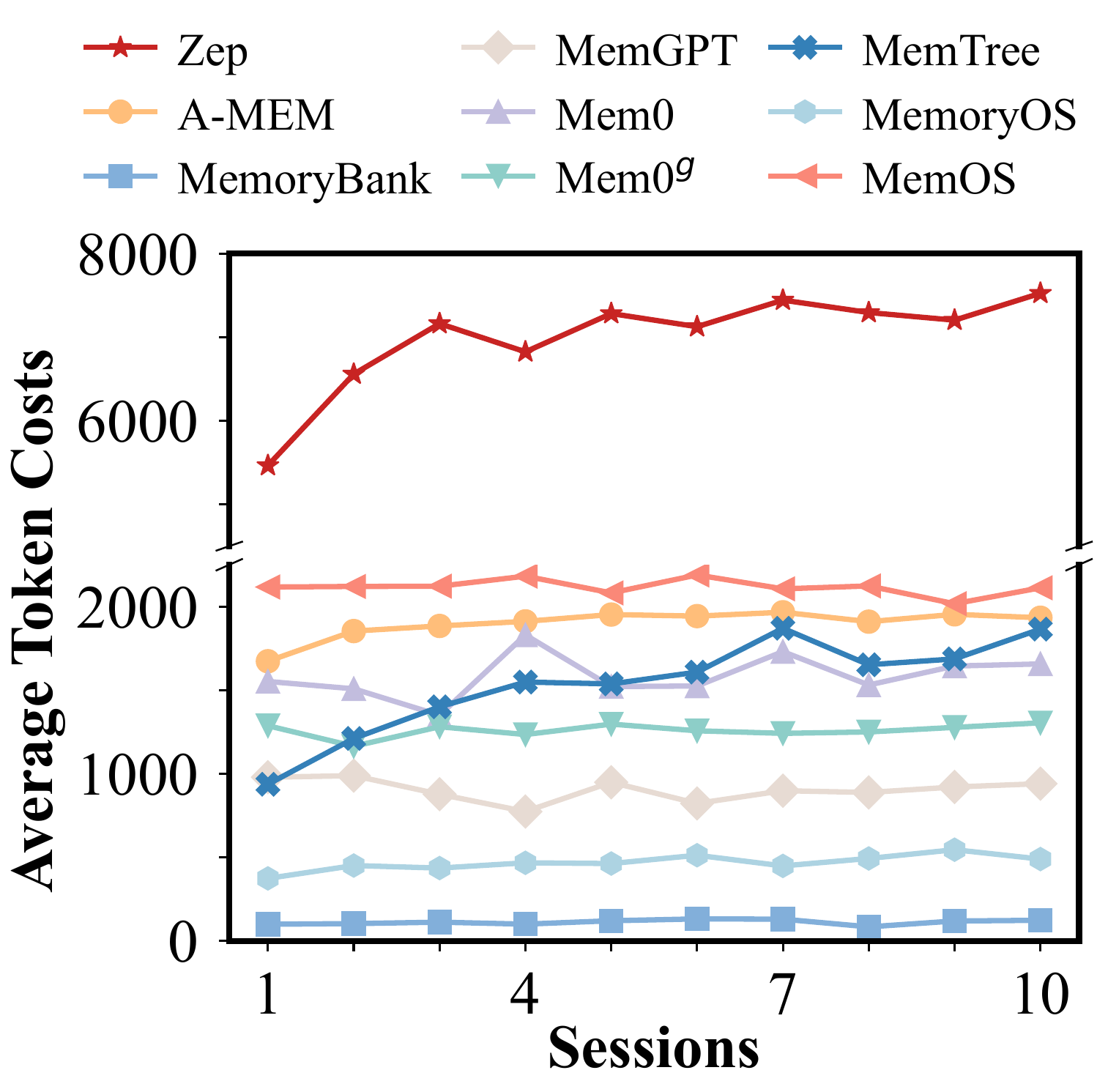}
        \caption{Average token costs}
        \label{fig:token}
    \end{subfigure}

    \caption{Performance and efficiency analysis on LOCOMO.}
    \label{fig:efficiency}
\end{figure}



\begin{figure*}[t]
    \centering
    
    \begin{subfigure}[]{0.88\linewidth}
        \centering
        \includegraphics[width=\linewidth]{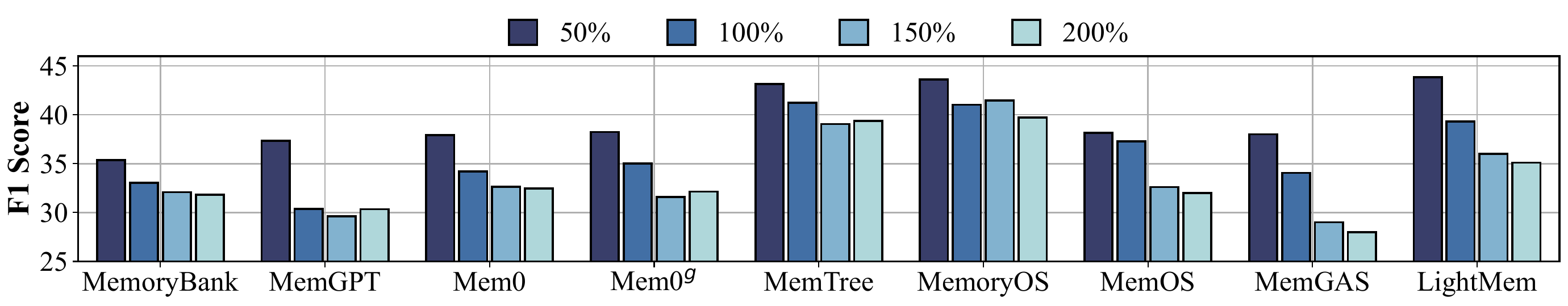}
        \caption{Context scalability}
        \label{fig:scale_overall}
    \end{subfigure}
    
    \begin{subfigure}[]{0.88\linewidth}
        \centering
        \includegraphics[width=\linewidth]{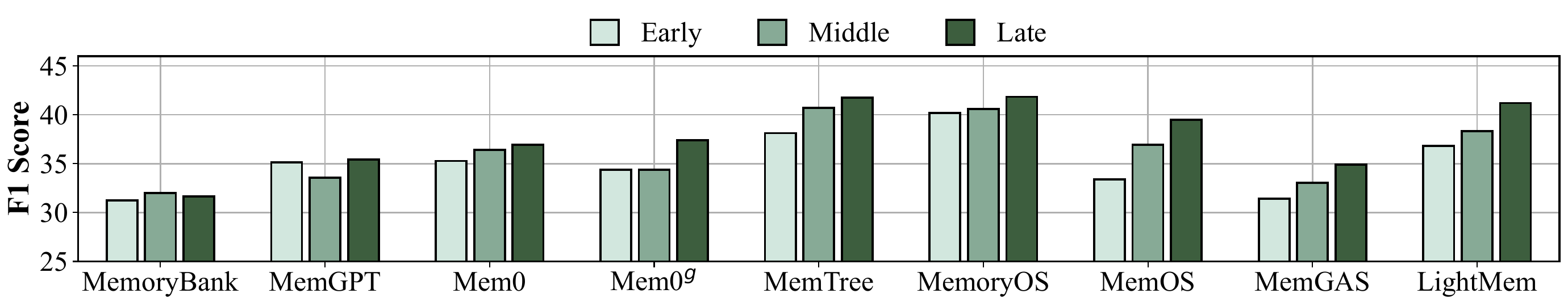}
        \caption{Position sensitivity}
        \label{fig:position_overall}
    \end{subfigure}

    \caption{\revise{Robustness analysis of memory mechanisms on LONGMEMEVAL. (a) illustrates the context scalability as the input scale varies from 50\% to 200\%, while (b) shows the position sensitivity regarding ground-truth information placed at different relative positions: Early (first 1/3), Middle (middle 1/3), and Late (last 1/3).}}
    \label{fig:robustness_overall}
\end{figure*}


\begin{table*}[]
\centering
\caption{\revise{Average retrieval latency per query (ms).}}
\label{tab:latency}
\setlength{\abovecaptionskip}{0pt}
\small
\revise{
\begin{tabular}{lrrrrrrrrrrrr}
\toprule
& \multicolumn{9}{c}{\textbf{Without Query-Time LLM Calls}}
& \multicolumn{3}{c}{\textbf{With Query-Time LLM Calls}} \\
\cmidrule(lr){2-10}\cmidrule(lr){11-13}
\textbf{Method} & \texttt{A-MEM} & \texttt{MemoryBank} & \texttt{MemGPT} & \texttt{Mem0} & \texttt{Zep} & \texttt{MemTree} & \texttt{MemOS} & \texttt{MemGAS} & \texttt{LightMem} & \texttt{Mem0$^g$} & \texttt{MemoChat} & \texttt{MemoryOS} \\
\midrule
\textbf{Latency (ms)} & 617.1 & 864.8 & 105.8 & 356.4 & 536.8 & \textbf{24.5} & 346.7 & 793.4 & \underline{78.1} & 3783.1 & 1792.4 & 2329.6 \\
\bottomrule
\end{tabular}
}
\vspace{-2mm}
\end{table*}

\revise{\begin{tikzpicture}
\filldraw (0,0) -- (-0.15,0.08) -- (-0.15,-0.08) -- cycle ; 
\end{tikzpicture} \textbf{Exp.2. Token cost and retrieval latency analysis.}
In this experiment, we evaluate the efficiency of each method during both memory construction and retrieval. For memory construction, we report token consumption, rather than runtime, as a proxy for computational cost, since this stage is dominated by LLM calls and its runtime is highly sensitive to deployment environments and system conditions. 
Specifically, we analyze: (1) the overall trade-off between performance and token cost, where Figure~\ref{fig:tradeoff} plots the average token cost per dialogue against the overall F1 score; and (2) construction scalability, where Figure~\ref{fig:token} shows how the average token consumption evolves as the volume of ingested memory increases. 
For retrieval, Table~\ref{tab:latency} reports the average elapsed time from when a memory method receives a query until the retrieval process is completed, before response generation begins. 
}

In general, higher performance correlates with increased token consumption, reflecting the benefit of extensive LLM utilization. 
\revise{While {\tt MemGAS} and {\tt MemOS} achieve high accuracy, they incur substantial token overhead. In contrast, {\tt MemoryOS} and {\tt LightMem} provide a better trade-off between performance and efficiency, achieving strong performance with significantly lower token costs.}
\revise{
As memory volume grows, certain methods, such as {\tt MemTree} and {\tt Zep}, exhibit poor token-cost scalability due to increasing memory update overheads.
}

\revise{
Retrieval latency varies substantially with the complexity of the retrieval pipeline. \texttt{MemTree} achieves the lowest retrieval latency (24.5\,ms), as it treats all tree nodes as a unified set and directly retrieves the relevant nodes without traversing the tree structure. \texttt{LightMem}, a representative lightweight memory method, achieves the second-lowest latency (78.1\,ms). In contrast, methods involving query-time LLM calls, for example, to extract entities from the query, generally incur substantially higher retrieval latency.
}

\begin{tikzpicture}
\filldraw (0,0) -- (-0.15,0.08) -- (-0.15,-0.08) -- cycle ; 
\end{tikzpicture} \textbf{Exp.3. Context scalability analysis.}\label{exp3}
In this experiment, we investigate the context scalability of various memory architectures by expanding the context length of LONGMEMEVAL from $50\%$ to $200\%$. In long-horizon interactions, the primary challenge for memory systems shifts from simple retrieval to robust noise suppression as information density grows. 
We evaluate various architectures across different context sizes to see how well they maintain retrieval precision as context grows.
Figure~\ref{fig:scale_overall} illustrates the overall trends in context scalability.
As the context scale expands from $50\%$ to $200\%$, nearly all memory architectures exhibit a steady decline in F1 scores. This performance attrition is primarily driven by the increased density of irrelevant information, which lowers the signal-to-noise ratio during retrieval. 
\revise{Further analysis of this experiment is provided in Appendix~\ref{sec:app:exp}.}

\begin{tikzpicture}
\filldraw (0,0) -- (-0.15,0.08) -- (-0.15,-0.08) -- cycle ; 
\end{tikzpicture} 
\textbf{Exp.4. Position sensitivity analysis.}
In this experiment, we evaluate how the placement of key evidence affects retrieval and reasoning on variants of LONGMEMEVAL by positioning the evidence in the early (first 1/3), middle (middle 1/3), or late (last 1/3) sections of the context. As evidence appears earlier in the context, memory systems must bridge larger temporal gaps and handle increased interference from subsequent dialogue. This experiment tests whether architectures maintain uniform access to historical records or exhibit a bias toward recent inputs. 
\revise{As illustrated in Figure~\ref{fig:position_overall}, 
a clear recency bias is observed at the overall level as the temporal distance between supporting evidence and the query increases.}
\revise{Further analysis of this experiment is provided in Appendix~\ref{sec:app:exp}.}

\begin{tikzpicture}
\filldraw (0,0) -- (-0.15,0.08) -- (-0.15,-0.08) -- cycle ; 
\end{tikzpicture} 
\textbf{Exp.5. LLM backbone comparison.}
In this experiment, we evaluate several memory methods across different LLM backbones. Specifically, we select representative and high-performing methods from each design paradigm for this cross-backbone evaluation. We therefore exclude simpler baselines to focus on methods with non-trivial memory architectures. The results are presented in Table~\ref{tab:model}. 
\revise{Overall, most methods achieve their best performance with the closed-source GPT-5.4-mini backbone. Among open-source models, Qwen3.5-27B generally outperforms DeepSeek-V4-Flash across methods. Scaling Qwen3.5 from 9B to 27B produces substantial gains for all memory methods, as discussed in Exp.1, indicating that existing memory architectures remain heavily dependent on the backbone's reasoning capability.}

\begin{table}[]
    \setlength{\textfloatsep}{0cm}
    \centering
    \caption{\revise{Comparison across LLM backbones on LOCOMO.}}
    \label{tab:model}
    \small
    \setlength{\tabcolsep}{4pt}
    \begin{threeparttable}
    \revise{
        \begin{tabular}{l|l|r|r|r|r}
        \toprule
        \textbf{Model} & Metric & \multicolumn{1}{c|}{\texttt{Zep}} & \multicolumn{1}{c|}{\texttt{MemoryOS}} & \multicolumn{1}{c|}{\texttt{MemOS}} & \multicolumn{1}{c}{\texttt{LightMem}} \\
        \midrule
        \multirow{2}{*}{Qwen3.5-9B}
          & F1    & 42.89 & 37.36 & 42.21 & 34.70 \\
          & BLEU-1 & 36.94 & 29.92 & 36.65 & 28.18 \\
        \midrule
        \multirow{2}{*}{Qwen3.5-27B}
          & F1    & 46.62 & 44.36 & 48.88 & 39.67 \\
          & BLEU-1 & 39.88 & 36.92 & 41.35 & 32.55 \\
        \midrule
        \multirow{2}{*}{DeepSeek-V4-Flash}
          & F1    & 43.16 & 42.82 & 48.52 & 38.88 \\
          & BLEU-1 & 38.71 & 35.31 & 40.11 & 31.69 \\
        \midrule
        \multirow{2}{*}{GPT-5.4-mini}
          & F1    & 50.10 & 46.66 & 50.93 & 40.64 \\
          & BLEU-1 & 41.69 & 38.85 & 42.45 & 32.72 \\
        \bottomrule
        \end{tabular}
    }
    \end{threeparttable}
\end{table}



\begin{figure}[]
    \centering
    \includegraphics[width=\linewidth]{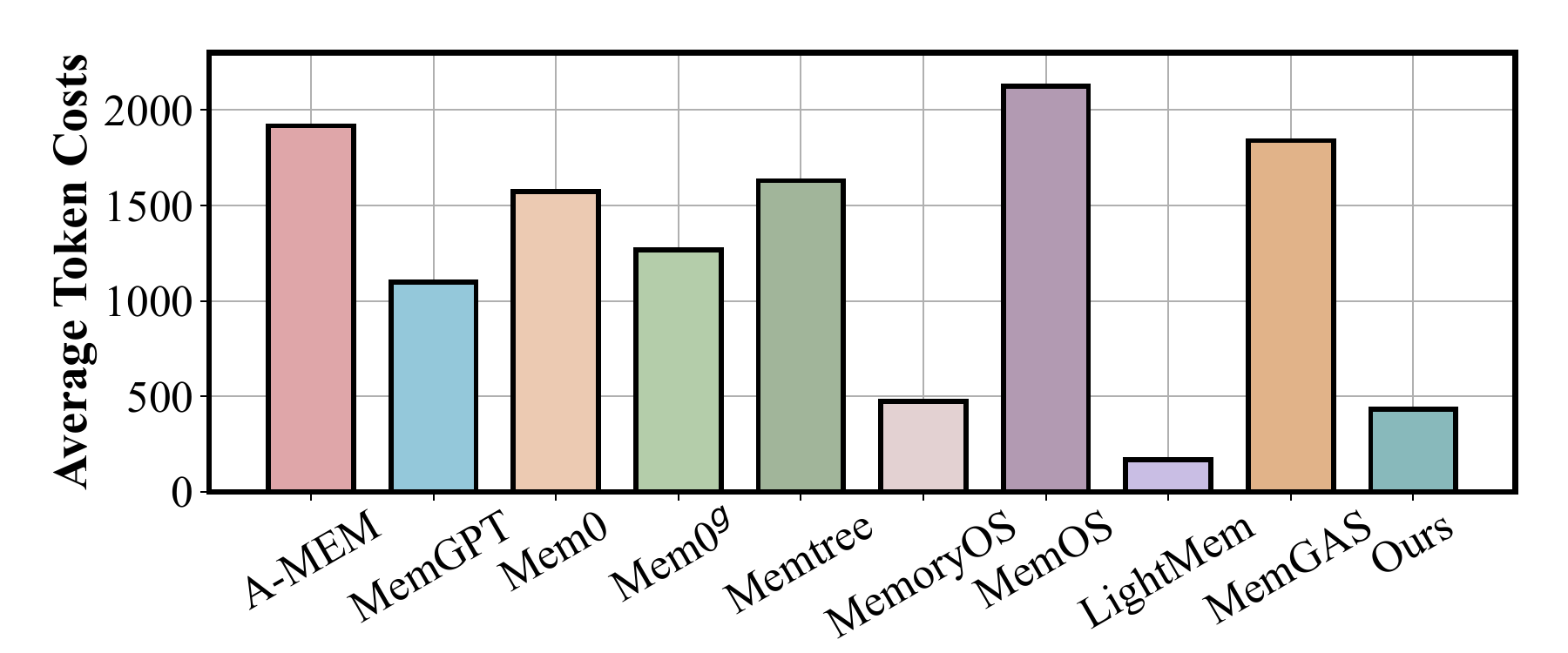}
    \setlength{\abovecaptionskip}{0pt}
    \setlength{\belowcaptionskip}{5pt}
    \caption{\revise{Comparison of our newly designed method in terms of average token costs.}}
    \label{fig:sota_cost}
\end{figure}

\begin{table}[]
\centering
\caption{\revise{Comparison of our newly designed method on LOCOMO and LONGMEMEVAL.}}
\label{tab:ours_result}
\small
\revise{
\begin{tabular}{lcccc}
\toprule
& \multicolumn{2}{c}{\textbf{LOCOMO}}
& \multicolumn{2}{c}{\textbf{LONGMEMEVAL}} \\
\cmidrule(lr){2-3}
\cmidrule(lr){4-5}
\textbf{Method} & F1 & BLEU-1 & F1
& BLEU-1 \\
\midrule

\tt{Zep}
& \underline{42.89} & \underline{36.94} & --- & --- \\

\tt{MemTree}
& 38.53 & 31.01 & \underline{41.24} & 34.69\\

\tt{MemoryOS}
& 37.36 & 29.92 & 41.02 & \underline{35.61}\\

\tt{MemOS}
& 42.21 & 36.65 & 37.29 & 30.66\\

\tt{LightMem}
& 34.70 & 28.18 & 39.31 & 32.49\\

\tt{Ours}
& \textbf{43.40} & \textbf{37.36} & \textbf{46.84} & \textbf{39.83}\\

\bottomrule
\end{tabular}
}
\end{table}

\begin{table}[]
\centering
\caption{
\revise{Component ablations and organization variants measured by F1 score on LOCOMO.}
}
\label{tab:ablation}
\small
\renewcommand{\arraystretch}{1.1}
\revise{
\begin{tabular}{lrrrrr}
\toprule
\textbf{Method Variants}
& \textbf{SH}
& \textbf{MH}
& \textbf{TR}
& \textbf{OK}
& \textbf{Overall} \\
\midrule
\textbf{Ours}
& 49.63 & 36.94 & 39.86 & 19.69 & 43.40 \\

\quad w/o Tree
& 45.82 & 35.23 & 31.89 & 14.25 & 39.01 \\

\quad w/o SP
& 47.97 & 34.99 & 35.24 & 19.20 & 41.15 \\

\quad w/o HM \& SP (\texttt{MemTree})
& 43.42 & 34.21 & 33.08 & 26.54 & 38.53 \\

\quad w/o Tree \& SP (\texttt{MemoryOS})
& 43.58 & 35.64 & 28.05 & 18.98 & 37.36 \\
\midrule
\quad w/o Tree, w/ Graph
& 45.52 & 33.68 & 40.06 & 19.59 & 40.60 \\

\quad w/o Tree \& SP, w/ Graph
& 45.97 & 26.84 & 40.88 & 21.59 & 39.89 \\
\bottomrule
\end{tabular}
}
\end{table}

\begin{table*}[t]
\centering
\caption{\revise{Comparison of methods on MemoryArena. The best and second-best results are marked in bold and underlined, respectively.}}
\small
\revise{
\begin{tabular}{lrrrrrrrrrrrr}
\toprule
\multicolumn{1}{l}{\multirow{3}[2]{*}{Method}} & \multicolumn{2}{c}{\multirow{2}[1]{2cm}{\centering Bundled Web Shopping}} & \multicolumn{3}{c}{\multirow{2}[1]{2.5cm}{\centering Group Travel Planning}} & \multicolumn{2}{c}{\multirow{2}[1]{2cm}{\centering Progressive Web Search}} & \multicolumn{4}{c}{\multirow{1}[1]{3.8cm}{\centering Formal Reasoning}} & \multicolumn{1}{c}{\multirow{3}[1]{1cm}{\centering All Task Avg SR}}\\ [-2pt] \cmidrule(lr){9-12}
 \multicolumn{8}{l}{} & \multicolumn{2}{c}{Math}  & \multicolumn{2}{c}{Phys} & \multicolumn{1}{c}{} \\ [-2pt] \cmidrule(lr){2-12}
\multicolumn{1}{r}{} & \multicolumn{1}{>{\centering\arraybackslash}m{0.8cm}}{SR} & \multicolumn{1}{r}{PS} & \multicolumn{1}{r}{SR} & \multicolumn{1}{r}{PS} & \multicolumn{1}{r}{sPS} & \multicolumn{1}{r}{Acc} & \multicolumn{1}{r}{\#Search} & \multicolumn{1}{r}{SR} & \multicolumn{1}{r}{PS} & \multicolumn{1}{r}{SR} & \multicolumn{1}{r}{PS} & \multicolumn{1}{r}{}\\ \midrule

{\tt A-MEM}
& 0.00 & 31.03
& 0.00 & \underline{0.27} & 6.32
& 2.26 & 7.04
& 10.00 & \underline{25.92}
& 35.00 & \textbf{56.54}
& 9.45 \\

{\tt MemoryBank}
& 0.00 & 32.08
& 0.00 & 0.21 & 5.97
& \textbf{5.43} & 6.09
& \underline{17.50} & 24.49
& 35.00 & \underline{54.04}
& 11.59 \\

{\tt MemGPT/Letta}
& 0.00 & \textbf{36.45}
& 0.00 & 0.00 & 4.82
& 3.62 & 7.09
& \textbf{20.00} & \textbf{28.58}
& 20.00 & 49.04
& 8.72 \\

{\tt Mem0}
& 0.00 & 31.91
& 0.00 & 0.00 & \underline{7.48}
& \textbf{5.43} & 7.78
& 10.00 & 23.49
& 30.00 & 49.18
& 9.09 \\

{\tt Mem0\(^g\)}
& 0.00 & \underline{32.50}
& 0.00 & 0.00 & 7.34
& \underline{4.98} & 7.60
& 12.50 & 24.47
& 10.00 & 46.97
& 5.50 \\

{\tt Zep}
& 0.00 & 27.56
& 0.00 & 0.05 & 6.60
& 2.26 & \textbf{4.49}
& 15.00 & 23.61
& \textbf{45.00} & 52.42
& \textbf{12.45} \\

{\tt MemTree}
& 0.00 & 28.89
& 0.00 & 0.16 & 6.29
& 2.26 & \underline{4.96}
& 12.50 & 24.54
& 30.00 & 46.22
& 8.95 \\

{\tt MemoryOS}
& 0.00 & 30.22
& 0.00 & \textbf{0.43} & \textbf{8.02}
& 2.26 & 8.13
& 15.00 & 21.80
& 15.00 & 42.51
& 6.45 \\

{\tt MemOS}
& 0.00 & 26.44
& 0.00 & 0.05 & 4.09
& 1.81 & 8.33
& 15.00 & 24.78
& 15.00 & 32.88
& 6.36 \\

{\tt MemGAS}
& 0.00 & 18.67
& 0.00 & 0.00 & 5.90
& 3.62 & 6.25
& 15.00 & 21.31
& \underline{40.00} & 45.92
& \underline{11.72} \\

{\tt LightMem}
& 0.00 & 30.11
& 0.00 & 0.00 & 0.08
& 1.36 & 10.51
& 7.50 & 20.93
& 25.00 & 35.26
& 6.77 \\
\bottomrule
\end{tabular}
}
\label{tab:arena}
\end{table*}

\begin{tikzpicture}
\filldraw (0,0) -- (-0.15,0.08) -- (-0.15,-0.08) -- cycle ; 
\end{tikzpicture} 
\textbf{Exp.6. New SOTA algorithm.}
\revise{
Based on the above analysis, we design a new memory framework that achieves state-of-the-art performance among the evaluated representative memory baselines under the same experimental settings while maintaining low token overhead. 
Motivated by their strong performance, we integrate the multi-granularity tree organization of \texttt{MemTree} and \texttt{MemOS} with the hierarchical memory storage of \texttt{MemoryOS}. To reduce the token overhead of tree updates, we further employ segment-level processing instead of processing individual dialogue turns. This design not only reduces update costs but also preserves semantic coherence, resulting in improved performance~\cite{xu2026memgas,pan2025secom}.
}
The detailed workflow and algorithm are provided in Appendix~\ref{sec:app:ours}.

\revise{Table~\ref{tab:ours_result} reports the performance metrics,} and Figure~\ref{fig:sota_cost} compares the average token costs. 
Our method achieves the best overall performance on both benchmarks and delivers highly competitive results across all task categories, while maintaining a remarkably low computational overhead of fewer than 450 tokens per dialogue. 

\revise{
We further conduct component-wise ablations on three key designs: tree-based organization (Tree), hierarchical memory separation (HM), and segment-level processing (SP), as reported in Table~\ref{tab:ablation}. Note that SH, MH, TR, and OK denote single-hop, multi-hop, temporal reasoning, and open-domain knowledge questions, respectively. Since SP operates during the transition from short-term to mid-term memory, removing HM necessarily removes SP. Removing Tree or SP reduces the overall F1 score by 4.39 and 2.25 points, respectively, while retaining only Tree (\texttt{MemTree}) or HM (\texttt{MemoryOS}) results in larger performance drops. These results demonstrate the contribution of all three components, with temporal reasoning being the most affected question category.

We also replace Tree with Graph while keeping the remaining framework unchanged, following \texttt{Zep}'s knowledge-graph construction pipeline. Replacing Tree with Graph reduces the overall F1 score from 43.40 to 40.60, showing that tree-based organization provides stronger and more balanced overall performance. Nevertheless, Graph performs better on temporal reasoning questions, suggesting that the temporal information retained in its nodes and edges is more effective for modeling temporal dependencies.
}

\revise{
\begin{tikzpicture}
\filldraw (0,0) -- (-0.15,0.08) -- (-0.15,-0.08) -- cycle ; 
\end{tikzpicture} 
\textbf{Exp.7. Sensitivity to retrieval top-$k$.}
We evaluate representative methods using different retrieval top-$k$ values, with $k\in\{1,3,5,10,15\}$, on LOCOMO, as shown in Figure~\ref{fig:topk}. Performance improves substantially as k increases from 1 to 10, while the gains become marginal beyond k=10. Based on this observation and following the default settings adopted in prior work~\cite{wujiang2025amem,rezazadeh2024isolated,kang2025memory}, we use k=10 as the default unless otherwise specified.
}

\begin{figure}[]
    \centering
    \setlength{\abovecaptionskip}{0cm}
    \includegraphics[width=0.78\linewidth]{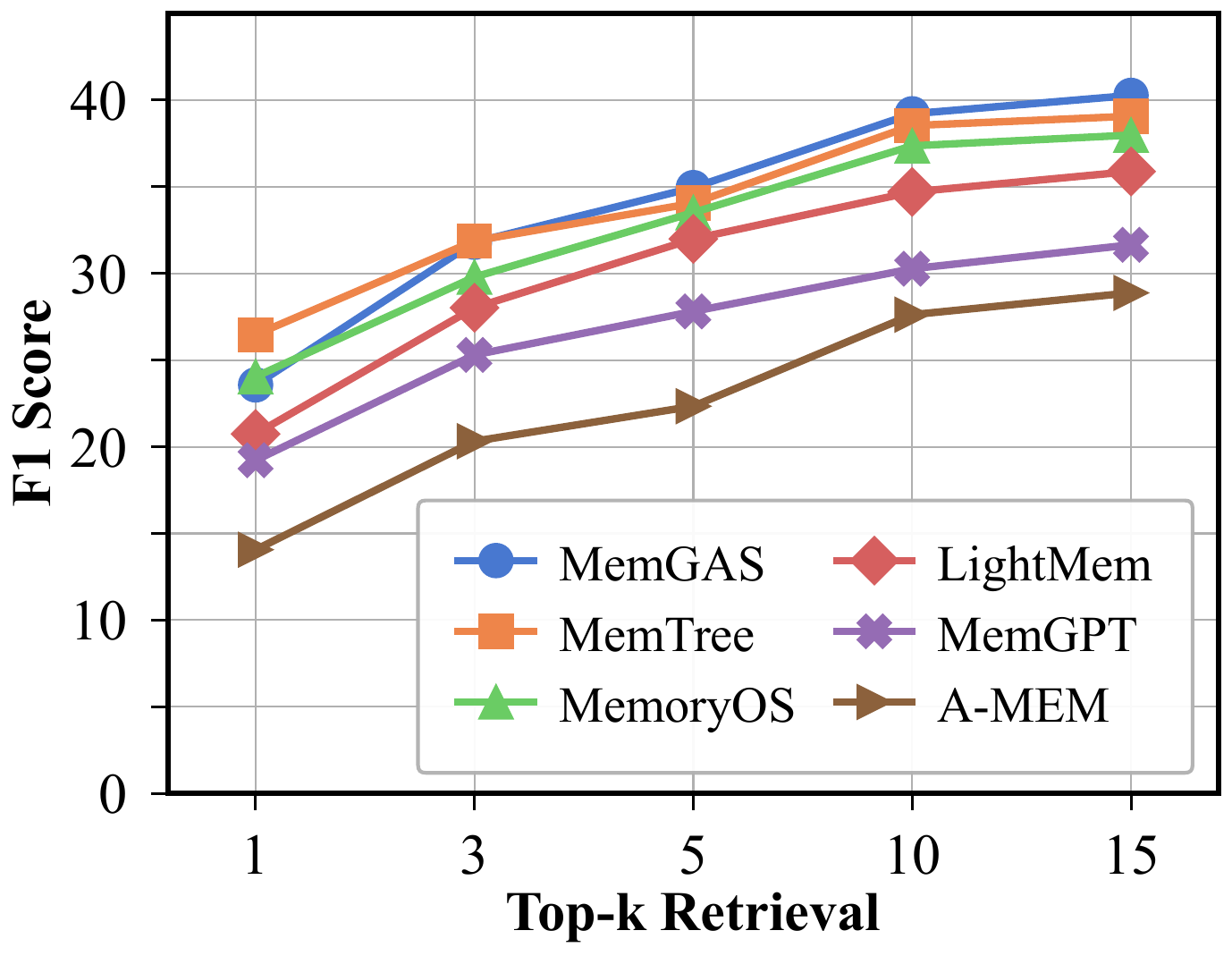}
    \caption{\revise{F1 performance of different memory methods under varying retrieval top-$k$ values.}}
    \label{fig:topk}
\end{figure}

\revise{
\begin{tikzpicture}
\filldraw (0,0) -- (-0.15,0.08) -- (-0.15,-0.08) -- cycle ; 
\end{tikzpicture} 
\textbf{Exp.8. Evaluation on Agentic Memory Tasks.}
In this experiment, we evaluate representative memory methods on MemoryArena, which measures memory-augmented agents in agent--environment interactions. The results are shown in Table~\ref{tab:arena}.
Overall, existing approaches still face substantial challenges in achieving reliable task completion.
Most methods achieve low SR scores, especially on \textit{Bundled Web Shopping} and \textit{Group Travel Planning}, where an SR of zero indicates that the agent fails to complete any full task trajectory. \texttt{Zep} achieves the best overall performance with an average SR of 12.45, followed by \texttt{MemGAS} and \texttt{MemoryBank}. The low success rates across tasks indicate that existing memory mechanisms are still insufficient for supporting long-horizon agentic interactions.
}

\section{Lessons and Opportunities}
\label{sec:lessons_opp}

\begin{table*}[]
\centering
\caption{\revise{Positioning of our work against representative benchmarks and surveys on agent memory.}}
\label{tab:position}

\setlength{\tabcolsep}{4pt}
\renewcommand{\arraystretch}{1.15}
\revise{
\resizebox{\textwidth}{!}{
\begin{tabular}{
    >{\raggedright\arraybackslash}p{8cm}
    cccccc
}
\toprule
\textbf{Work}
& \textbf{Primary Type}
& \textbf{Design Taxonomy}
& \textbf{\shortstack{Unified Multi-method\\Evaluation}}
& \textbf{\shortstack{Component--Performance\\Linkage}}
& \textbf{\shortstack{Evidence-guided\\Design Synthesis}} \\
\midrule

\textit{Memory in the Age of AI Agents}~\cite{hu2025memorysurvey}
& Survey
& \cmark
& \xmark
& \xmark
& \xmark \\

\textit{The AI Hippocampus: How Far Are We From Human Memory?}~\cite{jia2026hippocampus}
& Survey
& \cmark
& \xmark
& \xmark
& \xmark \\

\textit{A Survey on the Memory Mechanism of Large Language Model Based Agents}~\cite{zhang2025memorysurvey}
& Survey
& \cmark
& \xmark
& \xmark
& \xmark \\

\textit{Rethinking Memory Mechanisms of Foundation Agents in the Second Half}~\cite{huang2026rethinking}
& Survey
& \cmark
& \xmark
& \xmark
& \xmark \\

\midrule

\textit{LOCOMO}~\cite{maharana2024locomo}
& Benchmark
& \xmark
& \xmark
& \cmark
& \xmark \\

\textit{LongMemEval}~\cite{wu2025longmemeval}
& Benchmark
& \cmark
& \xmark
& \cmark
& \cmark \\

\textit{MemoryAgentBench}~\cite{hu2025memoryagentbench}
& Benchmark
& \xmark
& \cmark
& \xmark
& \xmark \\

\textit{MemoryArena}~\cite{he2026memoryarena}
& Benchmark
& \xmark
& \cmark
& \xmark
& \xmark \\

\midrule

\textbf{Memory in the LLM Era}
& Empirical Study
& \cmark
& \cmark
& \cmark
& \cmark \\

\bottomrule
\end{tabular}
}
}
\end{table*}

We summarize the lessons (L) for practitioners based on our observations and propose practical research opportunities (O).

\noindent \textbf{Lessons}:  

\revise{
\noindent \begin{tikzpicture}
\filldraw (0,0) -- (-0.15,0.08) -- (-0.15,-0.08) -- cycle ; 
\end{tikzpicture} \textbf{\underline{L1.}}
Based on the findings from all conducted experiments, we develop a roadmap in Figure~\ref{fig:roadmap} for selecting suitable memory architectures across four core capabilities: information persistence, memory association, temporal reasoning, and knowledge update. The detailed supporting analyses are provided in Appendix~\ref{sec:app:roadmap}.
}

\begin{figure}[]
    \centering
    \setlength{\abovecaptionskip}{0cm}
    \includegraphics[width=\linewidth]{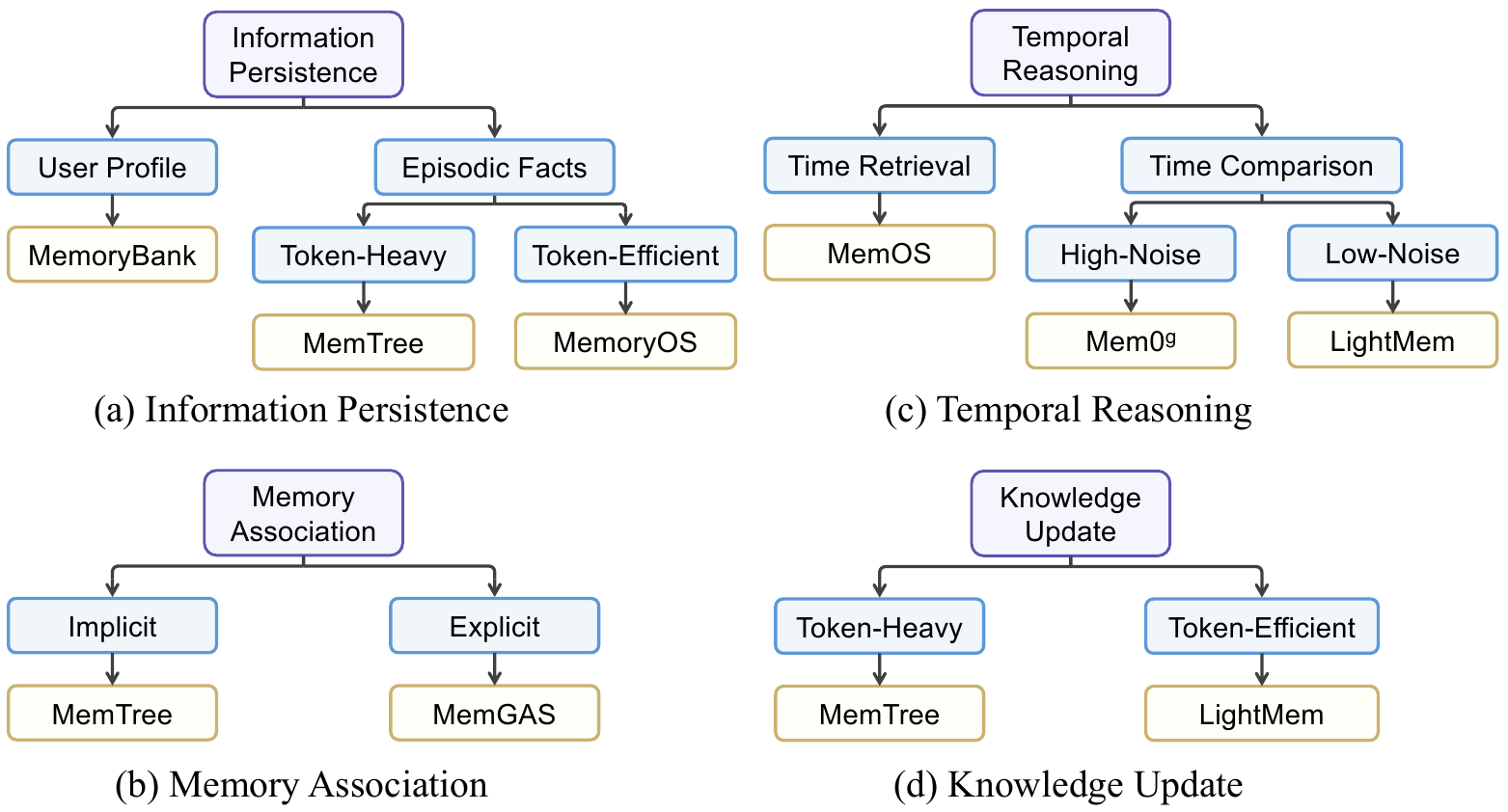}
    \caption{\revise{Task-Oriented Roadmap for Selecting Long-Term Memory Architectures under Different Scenarios.}}
    \label{fig:roadmap}
\end{figure}

\noindent \begin{tikzpicture}
\filldraw (0,0) -- (-0.15,0.08) -- (-0.15,-0.08) -- cycle ; 
\end{tikzpicture} \textbf{\underline{L2.}}
Compared to flat memory structures, hierarchical organization is more effective in capturing structural relationships between information, which can be achieved by either employing tree-based indices or designing multi-level storage.

\noindent \begin{tikzpicture}
\filldraw (0,0) -- (-0.15,0.08) -- (-0.15,-0.08) -- cycle ; 
\end{tikzpicture} \textbf{\underline{L3.}}
Information completeness is fundamental to memory mechanisms. While structured representations like triples in graphs improve organization, retaining raw dialogue context is essential to prevent semantic loss during the information extraction or retrieval stage.

\noindent \textbf{Opportunities}:

\noindent \begin{tikzpicture}
\filldraw (0,0) -- (-0.15,0.08) -- (-0.15,-0.08) -- cycle ; 
\end{tikzpicture} \textbf{\underline{O1.}}
In real-world settings, memory involves complex and diverse information sources, including textual conversations, historical interaction traces, and multimodal signals such as audio, images, and videos. Existing memory mechanisms often focus on a single or limited form, which constrains effective information utilization. A promising future research direction is to develop unified memory mechanisms that support heterogeneous and multimodal memory within a shared storage and retrieval framework.


\noindent \begin{tikzpicture}
\filldraw (0,0) -- (-0.15,0.08) -- (-0.15,-0.08) -- cycle ; 
\end{tikzpicture} \textbf{\underline{O2.}}
Existing competitive memory methods often manage and maintain information in ways that lead to rapid growth in storage size and increasing management and retrieval overhead. How to compress memory without losing useful information remains a major challenge, which creates an opportunity to explore latent representations beyond explicit text and learned compression mechanisms to achieve high-density yet usable memory.

\section{Related works}
\label{sec:related}

In this section, we briefly review the prior work most relevant to our study, including RAG frameworks and LLMs for database.

$\bullet$ \textbf{RAG frameworks.} RAG has been proven
to excel in many tasks, including open-ended question answering~\cite{jeong2024adaptive,siriwardhana2023improving}, programming context \cite{chen2024auto,chen2023haipipe,chen2024automatic}, SQL rewrite~\cite{lillm,sun2024r}, automatic DBMS configuration debugging~\cite{zhou2023d,singh2024panda}, and data cleaning~\cite{naeem2024retclean,narayan2022can,qian2024unidm}.
The naive RAG technique relies on retrieving query-relevant information from external knowledge bases to mitigate the ``hallucination'' of LLMs.
Recently, most RAG approaches~\cite{wu2024medical,wang2024knowledge,li2024dalk,gutierrez2024hipporag,edge2024local,guo2024lightrag,sarthi2024raptor,peng2024graph} have adopted graphs as the external knowledge to organize the information and relationships within documents, achieving improved overall retrieval performance. 
In terms of open-source software, a variety of graph databases are
supported by both the LangChain~\cite{Langchain} and LlamaIndex~\cite{llamaindex} libraries,
while a more general class of graph-based RAG applications is also emerging, including systems that
can create and reason over knowledge graphs in both Neo4j~\cite{neo4j} and NebulaGraph~\cite{nebula}. 
For more details, please refer to the recent surveys and experimental studies of graph-based RAG methods~\cite{peng2024graph,han2024retrieval,zhou2026indepth}.

$\bullet$ \textbf{LLMs for database.}  Due to the wealth of developer experience captured in a vast array of database forum discussions, recent studies \cite{zhou2023d,lao2023gptuner,fan2024combining,zhou2024db,lillm,sun2024r,chen2024automatic,giannankouris2024lambda,li2025agenttune,stolz2023galois} have begun leveraging LLMs to enhance database performance. 
For instance, GPTuner \cite{lao2023gptuner} proposes to enhance database knob
tuning using LLMs by leveraging domain knowledge to identify
important knobs and coarsely initialize their values for subsequent
refinement. 
Besides, D-Bot \cite{zhou2023d} proposes an LLM-based database
diagnosis system, which can retrieve relevant knowledge chunks
and tools, and use them to identify typical root causes accurately.
The LLM-based data analysis systems and tools have also been studied~\cite{liang2025revisiting,liu2025palimpchat,anderson2024design,lin2025twix,lin2024towards,patel2024lotus,liu2024declarative,chen2023seed,hong2025datainterpreter,li2023sheetcopilot,li2024autodcworkflow,zhang2024jellyfish,gao2024texttosql,yan2025mctuner}.

\revise{
$\bullet$ \textbf{Agent Memory and Data Management.} Agent memory provides the essential data management foundation for stateful LLM agents. 
It continuously ingests interaction streams, transforms them into logical memory records, maintains evolving and potentially conflicting information, organizes these records in textual, vector, hierarchical, or graph forms, and retrieves relevant context for incoming queries. 
Together, these operations constitute a data management pipeline for persistent agent states. 
This data management perspective is also reflected in recent database research~\cite{sun2025gaussdb,chang2025sagallm,khan2025rag,fernandez2023large}. 
Fernandez et al.~\cite{fernandez2023large} highlight the complementary role of databases and information retrieval systems in supplying and processing external information for LLMs.
GaussDB-Vector~\cite{sun2025gaussdb} explicitly treats a persistent vector database as long-term memory for LLM applications, supporting scalable search and real-time updates.
Khan et al.~\cite{khan2025rag} also identify data organization, indexing, retrieval, and updates in RAG as data management challenges.
}

\revise{
$\bullet$ \textbf{Agent Memory Surveys and Benchmarks.}
Recent surveys provide conceptual taxonomies of agent memory from different perspectives~\cite{hu2025memorysurvey,zhang2025memorysurvey,jia2026hippocampus,huang2026rethinking,yang2026graphbasedagentmemorytaxonomy,wu2025memorysurvey1,du2025rethinkingmemoryllmbased,jiang2026memorysurvey3,luo2026memorysurvey4,wu2026memorysurvey5}, while recent memory benchmarks (e.g., LOCOMO~\cite{maharana2024locomo}, LONGMEMEVAL~\cite{wu2025longmemeval}, MemoryAgentBench~\cite{hu2025memoryagentbench}, and MemoryArena~\cite{he2026memoryarena}) establish evaluation protocols for conversational and agentic memory capabilities. As summarized in Table~\ref{tab:position}, these studies primarily focus on conceptual organization or system-level evaluation. Our work complements them by decomposing representative memory methods into reusable memory modules, evaluating these methods under unified settings, and empirically connecting component-level design choices to performance differences. These findings further support evidence-guided memory design and architecture selection.
}





\section{Conclusions}
\label{sec:conclusions}

In this paper, we provide an in-depth experimental evaluation and comparison of existing memory methods. We first present a unified modular framework that abstracts memory mechanisms into four core memory modules—information extraction, memory management, memory storage, and information retrieval. Under this framework, \revise{we systematically evaluate representative memory methods on three benchmark datasets}, and further conduct multi-dimensional experiments to study token cost efficiency, context scalability, evidence position sensitivity and LLM backbone dependence. Based on experimental results and analyses, we develop a new memory variant by combining existing techniques, which achieves strong accuracy while maintaining low overhead. Finally, we summarize the lessons learned and propose practical research opportunities that can facilitate future studies.
    

\clearpage

\balance
\bibliographystyle{ACM-Reference-Format}
\bibliography{ref}

\newpage
\appendix

\section{Prompt Templates}
\label{sec:app:prompt}
This section details several prompt templates used in our evaluation. A sample prompt for graph-based information extraction is shown in Figure~\ref{fig:pmt:graph_extract}, \revise{the prompt for LLM-as-a-judge is detailed in Figure~\ref{fig:judgeprompt}}, and the prompt for post-processing answer simplification is provided in Figure~\ref{fig:pmt:simplify}.

\begin{figure}[]
\centering
\setlength{\abovecaptionskip}{0cm}
\begin{AIbox}{Prompt for Graph-based Extraction.}
\underline{\bf Entity Extraction:} 
{
\begin{tcolorbox}[colback=gray!10, colframe=gray!10, boxrule=0pt, sharp corners, left=0pt, right=0pt]
Previous messages:

\texttt{\{previous messages\}}

Current message: 

\texttt{\{current message\}}

Given the above conversation, extract entity nodes from the current message that are explicitly or implicitly mentioned. Note that ALWAYS extract the speaker/actor as the first node and extract other significant entities, concepts, or actors mentioned in the current message.\\
\\
\# INSTRUCTIONS:\\
1. ALWAYS extract the speaker/actor as the first node. The speaker is the part before the colon in each line of dialogue.

2. Extract other significant entities, concepts, or actors mentioned in the current message.

3. DO NOT create nodes for relationships or actions.

4. DO NOT create nodes for temporal information like dates, times or years (these will be added to edges later).

5. Be as explicit as possible in your node names, using full names.

6. DO NOT extract entities mentioned only in the previous messages.
\end{tcolorbox}
}
\underline{\bf Relation Extraction:} 
{
\begin{tcolorbox}[colback=gray!10, colframe=gray!20, boxrule=0pt, left=0pt, sharp corners, right=0pt]
Previous messages:

\texttt{\{previous messages\}}

Current message: 

\texttt{\{current message\}}

Entities:

\texttt{\{entities\}}

Given the above messages and entities, extract all facts pertaining to the listed entities from the current message. Note that extract facts only between the provided entities and each fact should represent a clear relationship between two DISTINCT nodes.\\
\\
\# INSTRUCTIONS:\\
1. Extract facts only between the provided entities.

2. Each fact should represent a clear relationship between two DISTINCT nodes.

3. The relation\_type should be a concise, all-caps description of the fact (e.g., LOVES, IS\_FRIENDS\_WITH,
WORKS\_FOR).

4. Provide a more detailed fact containing all relevant information.

5. Consider temporal aspects of relationships when relevant.
\end{tcolorbox}
}
\end{AIbox} 
\caption{A sample prompt for graph-based extraction.}
\label{fig:pmt:graph_extract}
\end{figure}

\section{Experiment Details}
\label{sec:app:exp}

\subsection{Dataset Details}
\label{sec:app:exp:detail}
\revise{\noindent In this section, we provide a detailed overview of the three benchmarks.}

\revise{The LOCOMO benchmark~\cite{maharana2024locomo} comprises ten long-term conversations for question-answering evaluation. Each conversation features an average of 198.6 questions spanning 27.2 sessions and approximately 588.2 dialogue turns between two speakers.}
The specific task categories are defined as follows:
\begin{itemize}[topsep=1mm, partopsep=0pt, itemsep=0pt, leftmargin=10pt]
    \item \textbf{Single-hop} questions require answers based on a single session.
    \item \textbf{Multi-hop} questions require synthesizing information from multiple different sessions.
    \item \textbf{Temporal reasoning} questions require temporal reasoning and capturing time-related data cues.
    \item \textbf{Open-domain knowledge} questions require integrating provided information with external knowledge such as commonsense or world facts.
\end{itemize}

\revise{The LONGMEMEVAL benchmark~\cite{wu2025longmemeval} contains 500 high-quality questions designed to evaluate four core long-term memory abilities.
Each question is grounded in a dedicated conversation history based on long-term user--AI interactions, averaging 50.2 sessions and approximately 115,000 tokens in length.
LONGMEMEVAL categorizes memory tasks to assess the following aspects:}

\begin{itemize}[topsep=1mm, partopsep=0pt, itemsep=0pt, leftmargin=10pt]
    \item \textbf{Information Extraction (IE):} Ability to recall specific information from extensive interactive histories, including the details mentioned by either the user (\textbf{single-session-user}) or the assistant (\textbf{single-session-assistant}), and whether the model can utilize the user information to generate a personalized response (\textbf{single-session-preference}).
    \item \textbf{Multi-Session Reasoning (MR):} Ability to synthesize the information across multiple history sessions to answer complex questions that involve aggregation and comparison.
    \item \textbf{Knowledge Updates (KU):} Ability to recognize the changes in the user’s personal information and update the knowledge of the user dynamically over time.
    \item \textbf{Temporal Reasoning (TR):} Awareness of the temporal aspects of user information, including both explicit time mentions and timestamp metadata in the interactions.
\end{itemize}

\revise{
The MemoryArena benchmark~\cite{he2026memoryarena} comprises multi-session agentic tasks designed to evaluate memory use in agent--environment interactions. Each task contains an average of 6.9 explicitly interdependent subtasks and approximately 57 agent actions. 
Tasks are categorized into four types: 
\begin{itemize}

    \item \textit{Bundled Web Shopping}: The agent sequentially purchases a bundle of related products in a web environment. Each later purchase must be compatible with previously selected items while satisfying additional constraints such as budget, price, or rating, thereby testing whether the agent can retain and reuse earlier decisions.

    \item \textit{Group Travel Planning}: Starting from a finalized itinerary for one traveler, additional travelers join incrementally with shared or personalized preferences. The agent must generate each new itinerary by referring to previous travelers’ plans and resolving cross-traveler constraints, such as joining an existing activity or selecting an option relative to another traveler’s choice.

    \item \textit{Progressive Web Search}: A complex search problem is decomposed into a sequence of dependent subqueries, with each subquery introducing an additional condition. The agent must preserve and integrate results from earlier searches so that the final answer satisfies all previously introduced constraints.

    \item \textit{Sequential Formal Reasoning}: Research-level mathematics and physics problems are decomposed into ordered chains of intermediate questions. Later questions depend on definitions, lemmas, and intermediate results established in earlier sessions, requiring the agent to retrieve and reuse prior reasoning correctly.
    
\end{itemize}
\noindent It therefore evaluates several complementary aspects of agent memory, including cross-session state retention, tool-assisted interaction, long-horizon constraint tracking, and the reuse of previously acquired information or reasoning results.

For implementation, we followed the official MemoryArena source code and evaluation workflow, while using Qwen3.5-9B as the LLM backbone.
We mainly report \textbf{Task Success Rate (SR)} and \textbf{Task Progress Score (PS)}. Given $N$ tasks, where task $S_i$ contains $|S_i|$ ordered subtasks and $|S_i^{\mathrm{pass}}|$ of them are successfully completed, the metrics are defined as:
\[
\mathrm{SR}
=
\frac{1}{N}
\sum_{i=1}^{N}
\mathbb{I}(S_i \text{ is successfully completed}),
\]
and
\[
\mathrm{PS}
=
\frac{1}{N}
\sum_{i=1}^{N}
\frac{|S_i^{\mathrm{pass}}|}{|S_i|}.
\]
SR measures end-to-end task completion, while PS provides a finer-grained measure of partial progress across interdependent subtasks.
For \textit{Group Travel Planning}, we additionally report the soft Progress Score (sPS), where each subtask receives partial credit based on the fraction of constraints it satisfies. The subtask scores are first averaged within each task and then averaged across all tasks.
For \textit{Progressive Web Search}, we report Accuracy, measured by the correctness of the final search query in each task, and \#Search, measured by the number of search-tool calls used to answer the final integrated query.
}

\begin{figure}[]
\centering
\setlength{\abovecaptionskip}{0cm}
\begin{AIbox}{Prompt for Answer Simplification.}
{
\begin{tcolorbox}[colback=gray!10, colframe=gray!10, boxrule=0pt, sharp corners, left=0pt, right=0pt]
Your task is to act as an answer simplifier. I will give you a question and a full-sentence answer. You must reduce the answer to its most critical component.\\
Follow these rules:\\
1.  **Extract the Core Information:** Identify the primary piece of information that directly answers the question.\\
2.  **Remove Extraneous Phrases:** Eliminate phrases like "Based on the information provided...", "The answer is...", "As per the document...", etc.\\
3.  **Omit Explanations:** Do not include any justifications, reasoning, or additional context from the original answer.\\
4.  **Be Concise:** The output should be the shortest possible string that still accurately answers the question.\\
\\
Example:\\
- Question: "What degree did I graduate with?"\\
- Original Answer: "Based on the information provided, you graduated with a degree in Business Administration."\\
- Simplified Answer: "Business Administration"\\
    \\
Here is the question and answer to simplify:\\
- Question: \texttt{\{question\}}\\
- Original Answer: \texttt{\{answer\}}\\
- Simplified Answer:
\end{tcolorbox}
}
\end{AIbox} 
\caption{Prompt for answer simplification.}
\label{fig:pmt:simplify}
\end{figure}

\begin{figure}[h]
    \centering
    \includegraphics[width=1.0\linewidth]{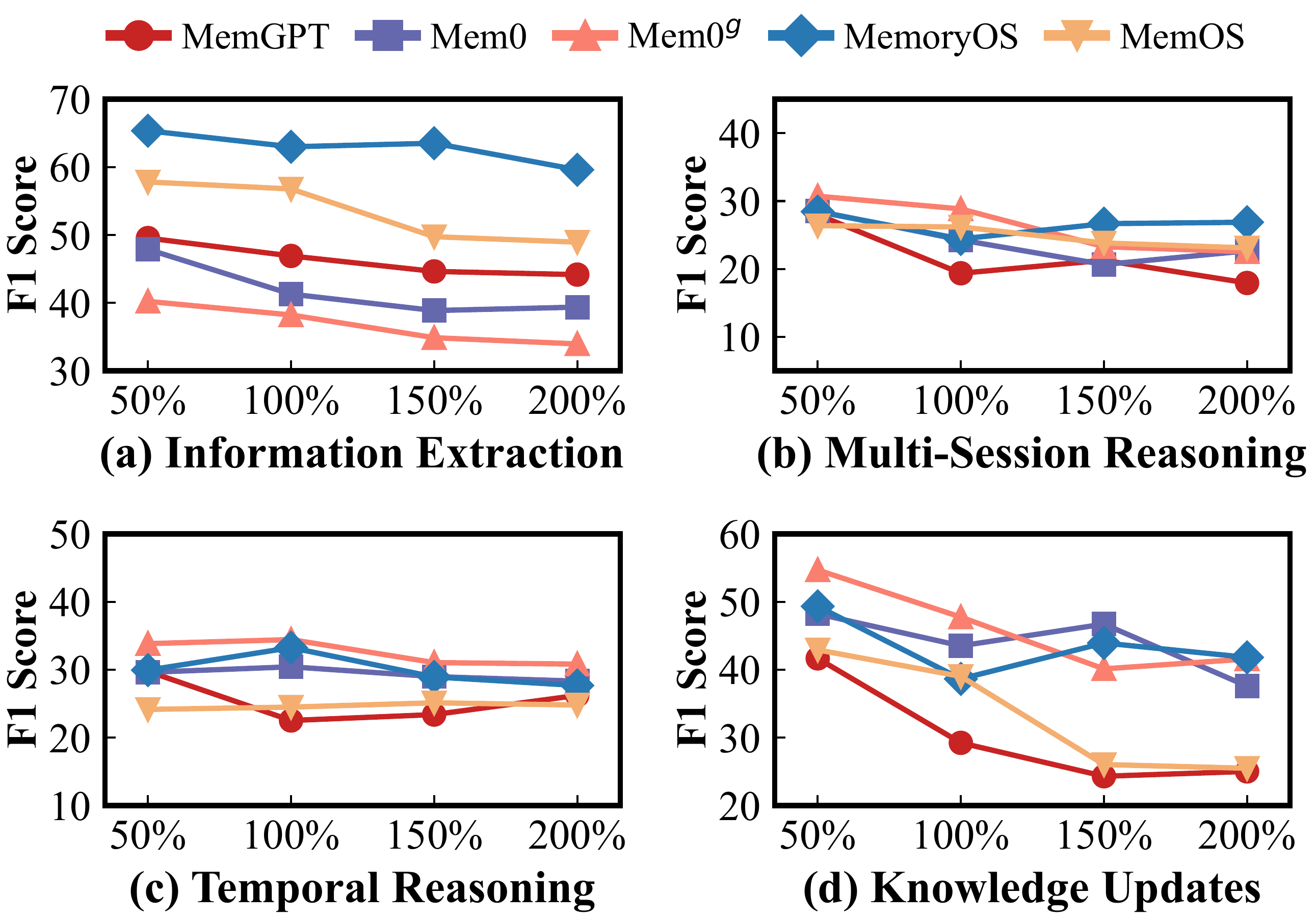}
    \caption{\revise{Context scalability of various memory methods across different task categories on LONGMEMEVAL.}}
    \label{fig:scale_tasks}
\end{figure}

\subsection{Dataset Variant Construction}
\label{sec:app:exp:construct}

\noindent LONGMEMEVAL is well-suited for controlled evaluation: it features a configurable context length enabled by a \emph{session pool}, which supplies topically similar history sessions without introducing conflicting information, and each sample is associated with a single question, making the corresponding ground-truth evidence clearly defined. Based on these properties, we construct a set of variants of the original dataset (LONGMEMEVAL$_S$) to evaluate \textbf{context scalability} and \textbf{position sensitivity}.

For \textbf{context scalability}, we construct variants at 50\%, 150\%, and 200\% of the default context length. Specifically, for the 50\% variant, we prune half of the history dialogues at session-level granularity while retaining the ground-truth session; for the 150\% and 200\% variants, we append additional sessions sampled from the session pool to the end of the original conversation history.

For \textbf{position sensitivity}, we generate position-varying variants of LONGMEMEVAL$_S$ by relocating the ground-truth session. Specifically, we extract the ground-truth session from its original position and re-insert it into one of three equally partitioned segments of the conversation: the first 1/3 (\emph{Early}), the middle 1/3 (\emph{Middle}), and the final 1/3 (\emph{Late}).

\begin{figure}[]
\centering
\setlength{\abovecaptionskip}{0cm}
\begin{AIbox}{Prompt for LLM-as-a-judge.}
{
\begin{tcolorbox}[colback=gray!10, colframe=gray!10, boxrule=0pt, sharp corners, left=0pt, right=0pt]
Your task is to label an answer to a question as ``{\tt CORRECT}'' or ``{\tt WRONG}''. You will be given the following data: (1) a question (posed by one user to another user), (2) a ‘gold’ (ground truth) answer, (3) a generated answer which you will score as {\tt CORRECT}/{\tt WRONG}.\\
\\
The point of the question is to ask about something one user should know about the other user based on their prior conversations. The gold answer will usually be a concise and short answer that includes the referenced topic, for example:\\
Question: Do you remember what I got the last time I went to Hawaii?\\
Gold answer: A shell necklace\\
The generated answer might be much longer, but you should be generous with your grading --- as long as it touches on the same topic as the gold answer, it should be counted as {\tt CORRECT}.\\
\\
For time-related questions, the gold answer will be a specific date, month, year, etc. The generated answer might be much longer or use relative time references (like ‘last Tuesday’ or ‘next month’), but you should be generous with your grading - as long as it refers to the same date or time period as the gold answer, it should be counted as {\tt CORRECT}. Even if the format differs (e.g., ‘May 7th’ vs ‘7 May’), consider it {\tt CORRECT} if it’s the same date.\\
\\
Now it’s time for the real question:\\
Question: \{question\}\\
Gold answer: \{gold\_answer\}\\
Generated answer: \{generated\_answer\}\\
\\
First, provide a short (one sentence) explanation of your reasoning, then finish with {\tt CORRECT} or {\tt WRONG}. Do NOT include both {\tt CORRECT} and {\tt WRONG} in your response, or it will break the evaluation script.\\
\\
Just return the label {\tt CORRECT} or {\tt WRONG} in a json format with the key as ``label''.
\end{tcolorbox}
}
\end{AIbox} 
\caption{\revise{Prompt for LLM-as-a-judge.}}
\label{fig:judgeprompt}
\end{figure}

\revise{
\subsection{Extended Context Scalability Analysis}

This section provides an extended analysis of context scalability under increasing context lengths.

The scaling results reveal a performance divide rooted in operational complexity. Methods such as {\tt MemOS} and {\tt MemGPT} implement an ``LLM-as-OS'' paradigm, requiring the LLM to autonomously manage memory via complex tool calls. As the scale increases to $200\%$, the expanded candidate space significantly raises the difficulty for the LLM to accurately reason over and execute these management instructions, leading to higher rates of tool-call failures and indexing conflicts. In contrast, {\tt MemoryOS} reduces the agent's cognitive load by employing explicit rule-based hierarchical management. By offloading organizational logic from the LLM to a deterministic framework, {\tt MemoryOS} maintains high stability, demonstrating that simplifying the agent's internal management overhead through deliberate design is critical for robust scaling.

Different task categories exhibit varying sensitivity to scaling pressure. As illustrated in Figure~\ref{fig:scale_tasks}, Knowledge Update (KU) is particularly sensitive, showing sharp performance attrition because an increased memory volume raises the density of conflicting records. Since KU requires the model to identify the latest fact among mutually exclusive versions, the presence of more obsolete candidates directly increases retrieval interference. Conversely, temporal tasks remain relatively stable, as they rely on the relative ordering of events, which remains structurally distinct even as the background volume grows. Unlike the versioning conflicts in KU, the chronological precedence of events is not easily compromised by the addition of context.
}

\begin{table}[h]
\centering
\caption{\revise{Position sensitivity analysis of representative memory mechanisms across Information Extraction sub-tasks on LONGMEMEVAL (F1 Score).}}
\small
\revise{
\begin{tabular}{llrrr}
\toprule

{\textbf{Method}} & {\textbf{Position}} & \textbf{user} & \textbf{assistant} & \textbf{preference} \\
\midrule

\multicolumn{1}{l}{\multirow{4}{*}{\tt Mem0$^g$}} &Early&63.80&23.35&11.39\\
&Middle&57.60&20.39&12.87\\
&Late&74.60&20.74&12.90\\
\cline{2-5}
& Improvement & \textbf{{+10.80}} & \textbf{{-2.61}} & \textbf{{+1.51}} \\
\midrule

\multicolumn{1}{l}{\multirow{4}{*}{\tt MemTree}} &Early&59.01&55.42&10.42\\
&Middle&66.91&67.61&12.07\\
&Late&71.49&65.41&11.33\\
\cline{2-5}
& Improvement & \textbf{{+12.48}} & \textbf{{+9.99}} & \textbf{{+0.91}} \\
\midrule

\multicolumn{1}{l}{\multirow{4}{*}{\tt MemOS}} &Early&69.40&49.09&12.64\\
&Middle&73.81&56.78&13.72\\
&Late&79.85&58.66&13.12\\
\cline{2-5}
& Improvement & \textbf{{+10.45 }} & \textbf{{+9.56 }} & \textbf{{+0.48 }} \\
\bottomrule

\label{tab:pos_IE}
\end{tabular}
}
\end{table}

\revise{
\subsection{Extended Position Sensitivity Analysis}
This section provides an extended analysis of position sensitivity with key evidence placed at different positions.

A clear recency bias is observed at the overall level as the temporal distance between supporting evidence and the query increases. As shown in Figure~\ref{fig:position_overall}, most methods achieve higher overall F1 scores when the evidence is placed in late sessions rather than early sessions.
As more intervening dialogue accumulates after the relevant evidence appears, maintaining long-range information consistency and retrieval becomes increasingly challenging. 

Memory update policy significantly affects position sensitivity. For methods like \texttt{A-MEM}, dynamic memory revisions can overwrite earlier evidence, making early-session information vulnerable to later interactions. Hierarchical methods like {\tt MemTree} and {\tt MemOS} exhibit different mechanisms: although raw dialogue is retained at leaf nodes, updates to higher-level summaries amplify recent information's influence, increasing the Late–Early gap. In contrast, {\tt MemoryOS} preserves historical information more evenly through stage-wise transfers across memory levels. Earlier evidence remains relatively independent within each level rather than being repeatedly merged with later information. Consequently, later interactions do not directly reshape earlier evidence representations, explaining {\tt MemoryOS}'s smaller Late–Early gap.

Position sensitivity is highly category-dependent. We examine three representative methods in Table~\ref{tab:pos_IE} across information extraction sub-tasks. Transient, session-localized information exhibits stronger position sensitivity than persistent traits. The user and assistant extraction tasks show substantial Early-to-Late changes while preference extraction remains stable.
This indicates that position sensitivity correlates with information persistence: transient session-local details suffer more from later interference, while persistent preference traits are less affected by evidence relocation.
}

\begin{algorithm}[]
  \caption{Our Newly Designed Method}
  \label{alg:ours}
  \small
  \SetKwInOut{Input}{input}\SetKwInOut{Output}{output}
  \Input{Current query $q$, Short-term memory $\mathcal{M}_{S}$, Mid-term memory $\mathcal{M}_{M}$, Long-term memory tree $\mathcal{M}_{L}$, short-term capacity threshold $\tau$, heat threshold $\theta$, beam width $k$}
  \Output{Updated memories $(\mathcal{M}'_{S}, \mathcal{M}'_{M}, \mathcal{M}'_{L})$, Response $r$}

  \tcp{\textcolor{teal}{(1) Information Retrieval}}
  $\mathcal{C}_{S} \gets \mathcal{M}_{S}$ \tcp*[r]{retrieve entire short-term memory}
  $\mathcal{C}_{M\_flat} \gets$ {\tt VectorSearch}$(\mathcal{M}_{M}.\text{high-level-nodes}, q)$\;
  $\mathcal{C}_{M\_beam} \gets$ {\tt BeamSearch}$(\mathcal{M}_{M}.\text{tree}, q, k)$\;
  $\mathcal{C}_{L} \gets$ {\tt VectorSearch}$(\mathcal{M}_{L}, q)$\;
  $\mathcal{C} \gets \mathcal{C}_{S} \cup \mathcal{C}_{M\_flat} \cup \mathcal{C}_{M\_beam} \cup \mathcal{C}_{L}$ \tcp*[r]{aggregate retrieved context}

  \tcp{\textcolor{teal}{(2) Response Generation}}
  $r \gets$ {\tt LLM\_Generate}$(q, \mathcal{C})$\;

  \tcp{\textcolor{teal}{(3) Memory Ingestion}}
  $\mathcal{M}'_{S} \gets$ {\tt Enqueue}$(\mathcal{M}_{S}, q, r)$ \tcp*[r]{ingest new messages via short-term FIFO queue}
  
  \If{$|\mathcal{M}'_{S}| > \tau$}{
      $\mathcal{M}_{old} \gets$ {\tt DequeueOldestHalf}$(\mathcal{M}'_{S})$\;
      $\mathcal{S}_{new} \gets$ {\tt SegmentBySemanticSimilarity}$(\mathcal{M}_{old})$ \tcp*[r]{partition into segments}
      $\mathcal{M}'_{M} \gets$ {\tt UpdateMemoryTree}$(\mathcal{M}_{M}, \mathcal{S}_{new})$ \tcp*[r]{leaf: segment, parents: aggregated summary}
  } \Else {
      $\mathcal{M}'_{M} \gets \mathcal{M}_{M}$\;
  }

  $\mathcal{M}'_{L} \gets \mathcal{M}_{L}$\;
  \ForEach{leaf node $n \in \mathcal{M}'_{M}.\text{leaves}$}{
      $score \gets$ {\tt ComputeHeatScore}$(n.\text{freq}, n.\text{recency})$\;
      \If{$score > \theta$}{
          $\mathcal{M}'_{L} \gets \mathcal{M}'_{L} \cup \{n\}$ \tcp*[r]{promote high-heat segments}
      }
  }

  \Return{$(\mathcal{M}'_{S}, \mathcal{M}'_{M}, \mathcal{M}'_{L}, r)$}
\end{algorithm}

\section{Details of Our Newly Designed Method}
\label{sec:app:ours}

This section details the complete workflow of our newly designed method and the algorithm is provided in Algorithm~\ref{alg:ours}.

As illustrated in Figure~\ref{fig:sota}, new messages are first ingested into short-term memory and managed via a FIFO queue. When the short-term memory reaches capacity, the oldest messages are partitioned into segments based on semantic similarity and transferred to mid-term memory. Within this tier, we maintain a memory tree in which each leaf node represents a segment---capturing a generated summary of its constituent messages---while parent nodes provide aggregated summaries of their respective children. This segment-level granularity significantly reduces token overhead compared to turn-level processing. For each segment leaf node, we compute a \emph{heat score} based on access frequency and recency; segments with high heat scores are promoted to long-term memory.

During the information retrieval stage, we conduct independent retrieval from each of the three storage tiers. Short-term memory is retrieved in its entirety to maintain context continuity. For mid-term memory, we employ a dual-mode retrieval mechanism: high-level node semantics are matched via flat vector-based similarity search, while raw messages are retrieved through a beam search that traverses from the root node, selecting the top-$k$ most similar nodes at each level and ultimately reaching the raw messages stored as children of the segment leaf nodes. Long-term memory is accessed via standard vector-based similarity retrieval.

\begin{figure}[h]
    \centering
    \includegraphics[width=0.97\linewidth]{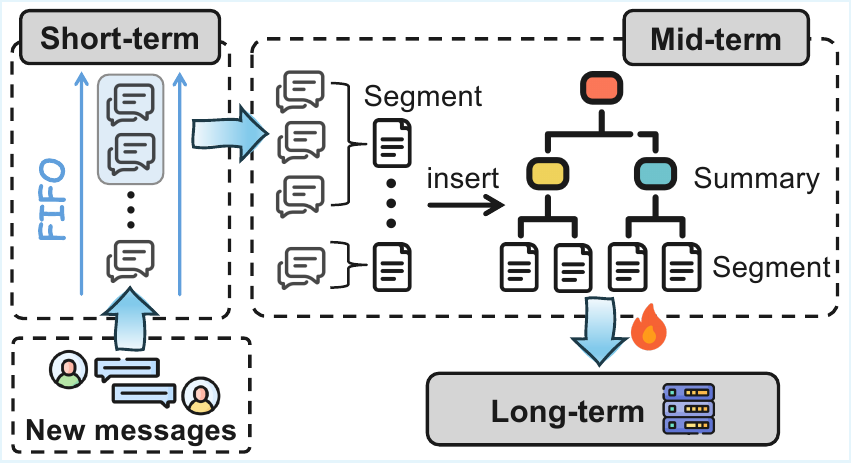}
    \setlength{\abovecaptionskip}{3pt}
    \setlength{\belowcaptionskip}{1pt}
    \caption{The framework of our newly designed method.}
    \label{fig:sota}
\end{figure}

\section{More Lessons and Opportunities}

\noindent \textbf{Lessons}: 

\noindent \begin{tikzpicture}
\filldraw (0,0) -- (-0.15,0.08) -- (-0.15,-0.08) -- cycle ; 
\end{tikzpicture} \textbf{\underline{L4.}}
Optimizing memory granularity by processing multiple dialogue turns as a single unit during the information extraction or memory management stage significantly reduces token consumption, while appropriate partitioning further maintains the coherence of retrieved information.

\noindent \begin{tikzpicture}
\filldraw (0,0) -- (-0.15,0.08) -- (-0.15,-0.08) -- cycle ; 
\end{tikzpicture} \textbf{\underline{L5.}}
\revise{
Well-designed memory frameworks can reduce the dependence of reasoning performance on model scale, enabling smaller LLMs to effectively handle complex queries such as those involving temporal dependencies.
}
For instance, employing multi-step reasoning or specialized components within the framework allows smaller LLMs to address complex tasks more effectively.

\noindent \begin{tikzpicture}
\filldraw (0,0) -- (-0.15,0.08) -- (-0.15,-0.08) -- cycle ; 
\end{tikzpicture} \textbf{\underline{L6.}}
Instead of destructive updates, memory systems should adopt non-destructive strategies that preserve historical information while annotating its validity, enabling future reuse and preventing the loss of potentially useful knowledge. In other words, instead of deleting old information, systems should keep it and mark its status.

\noindent \textbf{Opportunities}:

\noindent \begin{tikzpicture}
\filldraw (0,0) -- (-0.15,0.08) -- (-0.15,-0.08) -- cycle ; 
\end{tikzpicture} \textbf{\underline{O3.}}
Existing hierarchical memory mechanisms primarily focus on consolidating short-term memory into long-term storage but do not support the reverse transformation. A promising direction is to design bidirectional memory transformation mechanisms that enable efficient consolidation and reconstruction across memory hierarchies.

\noindent \begin{tikzpicture}
\filldraw (0,0) -- (-0.15,0.08) -- (-0.15,-0.08) -- cycle ; 
\end{tikzpicture} \textbf{\underline{O4.}}
In complex query scenarios, relying on a single fixed retrieval strategy can be brittle and may lead to task failure. Different queries often require different retrieval granularity and mechanisms. However, existing memory systems do not consider this issue, so designing a retrieval routing planner that dynamically selects and adapts retrieval strategies to diverse memory-query contexts is an interesting research problem.


\noindent \begin{tikzpicture}
\filldraw (0,0) -- (-0.15,0.08) -- (-0.15,-0.08) -- cycle ; 
\end{tikzpicture} \textbf{\underline{O5.}}
Existing memory benchmarks, such as LOCOMO and LONGMEMEVAL, evaluate memory methods on pre-collected, static interaction histories, which do not reflect the continuous and evolving nature of real-world memory. They also fail to capture key properties of human memory such as the preference for recent updates and the strengthening of repeatedly mentioned information. Moreover, despite the inclusion of sparse visual elements in LOCOMO, these benchmarks remain strictly text-centric. Developing more challenging, interaction-driven, and comprehensive multimodal benchmarks that better reflect realistic memory formation and usage scenarios is a meaningful research direction.

\begin{figure}[]
    \centering
    \setlength{\abovecaptionskip}{0cm}
    \includegraphics[width=0.9\linewidth]{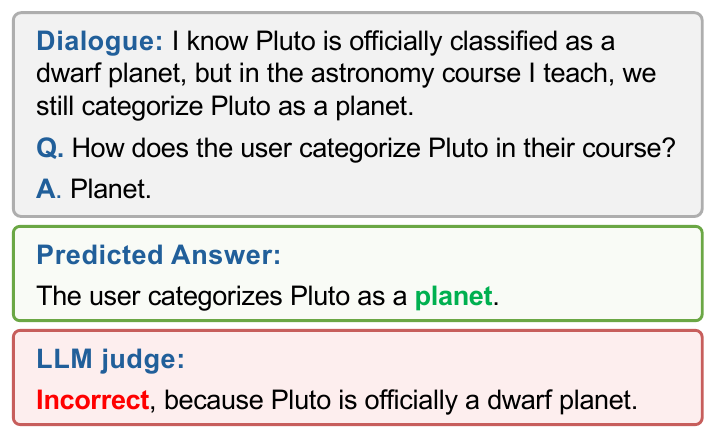}
    \caption{\revise{An example where a correctly generated answer is incorrectly evaluated by an LLM judge.}}
    \label{fig:judgecase}
\end{figure}

\revise{
\section{Analysis of Evaluation Metrics}
\label{sec:app:metric}

\subsection{LLM-as-a-judge}
It is worth noting that LLM-as-a-judge evaluation is not perfectly reliable, as its results may be affected by the capability of the judge model, the evaluation prompt, and other factors~\cite{zheng2023judging,li2025judge1,gu2024judge2,li2024judge3}. Figure~\ref{fig:judgecase} presents a concrete example of this issue. Therefore, we use it as a complementary metric rather than a primary evaluation metric.

\subsection{F1 and BLEU-1}
Token-level F1 measures the token overlap between a generated answer \(Y\) and its reference answer \(Y^{*}\). Specifically, precision, recall, and F1 are calculated as
\begin{equation}
P=\frac{\left|Y\cap Y^{*}\right|}{|Y|},
R=\frac{\left|Y\cap Y^{*}\right|}{|Y^{*}|},
\mathrm{F1}=\frac{2PR}{P+R},
\end{equation}
where \(\left|Y\cap Y^{*}\right|\) denotes the number of overlapping tokens, with repeated tokens counted according to their minimum occurrence count in the two answers. 
BLEU-1 measures clipped unigram precision with a brevity penalty and is calculated as
\begin{equation}
\mathrm{BLEU\mbox{-}1}
=
\mathrm{BP}\cdot p_{1},
\end{equation}
where
\begin{equation}
p_{1}
=
\frac{
\displaystyle\sum_{w}
\min\!\left(
\operatorname{Count}_{Y}(w),
\operatorname{Count}_{Y^{*}}(w)
\right)
}{
\displaystyle\sum_{w}\operatorname{Count}_{Y}(w)
},
\end{equation}
and the brevity penalty is defined as
\begin{equation}
\mathrm{BP}
=
\begin{cases}
1, & |Y|\geq |Y^{*}|,\\[2pt]
\exp\!\left(1-\dfrac{|Y^{*}|}{|Y|}\right), & |Y|<|Y^{*}|.
\end{cases}
\end{equation}
Here, \(w\) denotes a unigram token, and \(\operatorname{Count}_{Y}(w)\) and \(\operatorname{Count}_{Y^{*}}(w)\) denote the number of occurrences of \(w\) in the generated and reference answers, respectively.

Therefore, F1 and BLEU-1 are sensitive not only to whether the generated answer contains the correct information, but also to its wording and verbosity. Additional explanatory content may introduce unmatched tokens and reduce lexical precision even when the key answer is correct. This issue is particularly relevant because the official implementations of different memory methods use different response prompts, resulting in differences in answer length and style. To reduce this prompt-induced output-style bias and more accurately compare the information contained in the answers, we apply the same simplification procedure to the responses of all methods before computing F1 and BLEU-1. 
Table~\ref{tab:simplification_example} presents a concrete example illustrating how simplification affects both metrics, where removing semantically redundant details substantially increases the scores of the correct response and prevents it from being ranked below an incorrect but lexically closer answer. 
Table~\ref{tab:simplification_results} reports the F1 and BLEU-1 scores of different methods before and after simplification, thereby directly quantifying its impact on the evaluation results.
}

\begin{table}[]
\centering
\caption{\revise{Example of the effect of answer simplification on F1 and BLEU-1 scores.}}
\label{tab:simplification_example}
\small
\renewcommand{\arraystretch}{1.2}
\revise{
\begin{tabularx}{\linewidth}{
    >{\raggedright\arraybackslash}p{0.22\linewidth}
    >{\raggedright\arraybackslash}X
    >{\raggedleft\arraybackslash}p{0.05\linewidth}
    >{\raggedleft\arraybackslash}p{0.12\linewidth}
}
\toprule
\textbf{Question:}
& \multicolumn{3}{l}{When did Caroline go to the LGBTQ support group?} \\ [1pt]

\textbf{Answer:}
& \multicolumn{3}{l}{7 May 2023} \\ [1pt]
\toprule

\rowcolor{gray!20}
\textbf{Response type}
& \textbf{Predicted answer}
& \textbf{F1}
& \textbf{BLEU-1} \\
\midrule

\textcolor{green!60!black}{Correct} \textit{(before simplification)}
& Caroline went to the LGBTQ support group on May 7, 2023.
& 0.3750
& 0.2307 \\ [2pt]

\textcolor{green!60!black}{Correct} \textit{(after simplification)}
& May 7, 2023
& 0.8571
& 0.7500 \\ [2pt]

\textcolor{red!75!black}{Incorrect}
& 8 May 2023
& 0.6667
& 0.6667 \\
\bottomrule
\end{tabularx}
}
\end{table}

\begin{table}[]
\centering
\caption{\revise{F1 and BLEU-1 scores of different methods before and after answer simplification on LOCOMO.}}
\label{tab:simplification_results}
\small
\revise{
\begin{tabular}{lcccccc}
\toprule
& \multicolumn{3}{c}{\textbf{F1}}
& \multicolumn{3}{c}{\textbf{BLEU-1}} \\
\cmidrule(lr){2-4}
\cmidrule(lr){5-7}
\textbf{Method} & Before & After & $\Delta$
& Before & After & $\Delta$ \\
\midrule

\tt{A-MEM}
& 26.90 & 27.62 & +0.72 & 21.98 & 22.57 & +0.59 \\

\tt{Mem0}
& 31.40 & 34.55 & +3.15 & 25.32 & 28.78 & +3.80 \\

\tt{Zep}
& 35.93 & 42.89 & +6.96 & 30.78 & 36.94 & +6.16 \\

\tt{MemTree}
& 35.36 & 38.53 & +3.17 & 29.88 & 31.01 & +1.13 \\

\tt{MemoryOS}
& 35.72 & 37.36 & +1.64 & 26.15 & 29.92 & +3.77 \\

\tt{LightMem}
& 29.88 & 34.70 & +4.82 & 24.79 & 28.18 & +3.39 \\

\bottomrule
\end{tabular}
}
\end{table}

\begin{table}[]
\centering
\caption{\revise{Classification of the selected memory baselines by storage organization and memory representation.}}
\label{tab:storage_coverage}
\small
\setlength{\tabcolsep}{7pt}
\renewcommand{\arraystretch}{1.15}
\revise{
\begin{tabular}{lccc}
\toprule
\rowcolor{gray!20}
\textbf{Organization}
& \textbf{Vector}
& \textbf{Graph}
& \textbf{Tree} \\
\midrule
\textbf{Flat}
& \texttt{MemoryBank}
& \texttt{Mem0}$^{g}$
& \texttt{MemTree} \\
\textbf{Hierarchical}
& \texttt{MemoryOS}
& \texttt{Zep}
& \texttt{MemOS} \\
\bottomrule
\end{tabular}
}
\end{table}

\revise{
\section{Discussion of Method Selection}
\label{sec:app:method_selection}

In this section, we explain the coverage of the selected baselines from two complementary perspectives: the taxonomy in \textit{Memory in the Age of AI Agents: A Survey}~\cite{hu2025memorysurvey} and our four-stage framework.

\begin{figure}[]
    \centering
    \setlength{\abovecaptionskip}{0cm}
    \includegraphics[width=\linewidth]{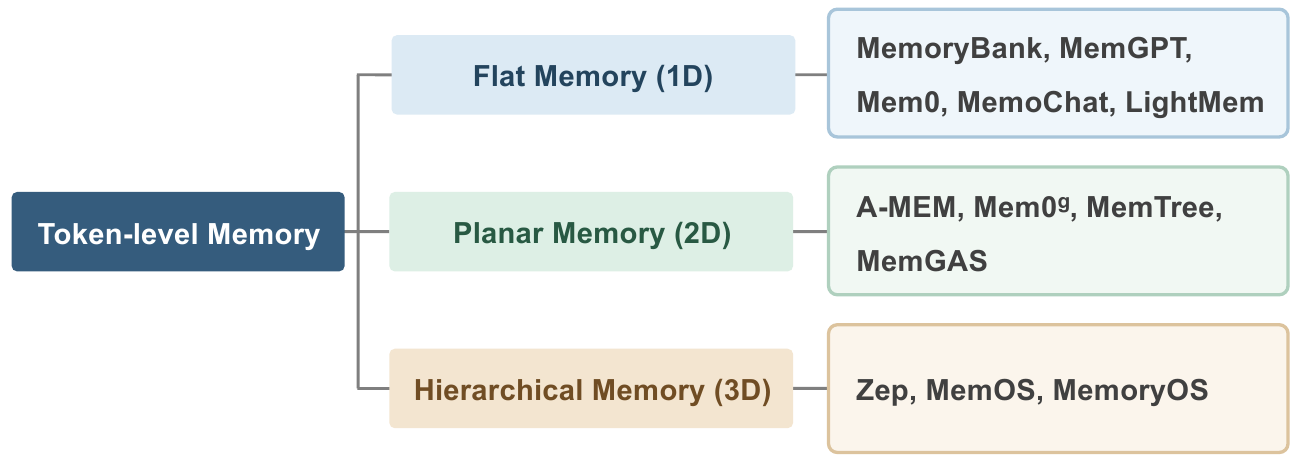}
    \caption{\revise{Classification and coverage of the evaluated token-level memory baselines following the taxonomy in \textit{Memory in the Age of AI Agents: A Survey.}}}
    \label{fig:survey_tax}
\end{figure}

Following \textit{Memory in the Age of AI Agents: A Survey}, which categorizes agent memory into token-level, parametric, and latent memory, we focus on text-based token-level memory. Unlike parametric and latent memory, these methods can be evaluated under the same frozen-LLM setting without additional training or model modification.
Within token-level memory, the survey further categorizes existing methods into Flat Memory (1D), Planar Memory (2D), and Hierarchical Memory (3D). 
As shown in Figure~\ref{fig:survey_tax}, for Flat Memory, we select representative methods covering classic conversational memory (e.g., {\tt MemoChat} and {\tt MemoryBank}), widely adopted designs (e.g., {\tt Mem0}), efficiency-oriented designs (e.g., {\tt LightMem}), and operating-system-inspired memory management (e.g., {\tt MemGPT}). 
For Planar Memory, we include associative memory (e.g., {\tt A-MEM}), knowledge-graph-based memory (e.g., {\tt Mem0g}), tree-structured memory (e.g., {\tt MemTree}), and multi-granularity graph memory (e.g., {\tt MemGAS}). 
For Hierarchical Memory, we select representative multi-level designs, including {\tt MemoryOS}, {\tt MemOS}, and the layered temporal graph architecture {\tt Zep}.

Beyond this survey-based taxonomy, the selected methods also cover diverse design choices across all four stages of our framework, as summarized in Table~\ref{tab:image_memory_methods}. 
Specifically, for \textit{information extraction}, the selected methods cover direct archiving at both the message level (e.g., {\tt MemoryBank}) and segment level (e.g., {\tt MemoryOS}), summarization-based extraction (e.g., {\tt A-MEM}), graph-based extraction (e.g., {\tt Zep} and {\tt Mem0g}), as well as the multi-granularity extraction design, {\tt MemGAS}. 
For \textit{memory management}, we include representative agentic approaches (e.g., {\tt Mem0} and {\tt LightMem}) and operating-system-inspired approaches (e.g., {\tt MemGPT} and {\tt MemOS}). 
For \textit{memory storage}, as shown in Table~\ref{tab:storage_coverage}, the selected methods cover representative combinations across both organization-centric and representation-centric dimensions.
For \textit{information retrieval}, they cover lexical-based, vector-based, and structure-based retrieval, as well as hybrid designs that combine multiple retrieval mechanisms, such as {\tt Zep} and {\tt MemGAS}.
}

\revise{
\section{Supporting Analysis for the Memory Architecture Selection Roadmap}
\label{sec:app:roadmap}
In this section, we provide the supporting analysis for the memory architecture selection roadmap. The roadmap covers four major capabilities of long-term memory systems: \textit{Information Persistence}, \textit{Memory Association}, \textit{Temporal Reasoning}, and \textit{Knowledge Update}.
These capabilities are derived from the major QA categories in widely used benchmarks such as LOCOMO and LONGMEMEVAL. They offer complementary perspectives on long-term memory requirements and collectively provide broad coverage of common QA scenarios. 
The method recommendations are primarily based on the comprehensive performance of different methods across LOCOMO and LONGMEMEVAL. When multiple methods achieve comparable effectiveness but are better suited to different scenarios, we provide scenario-specific recommendations by further considering factors such as token consumption and robustness under different noise levels.

For \textit{Information Persistence}, user-profile scenarios require memory mechanisms that model and maintain user preferences, as represented by \texttt{MemoryBank}. For episodic facts involving retrieval of specific information from historical interactions, \texttt{MemTree} is suitable when higher token overhead is acceptable, whereas the hierarchical design of \texttt{MemoryOS} provides a more token-efficient solution.

For \textit{Memory Association}, the roadmap distinguishes implicit from explicit associations. 
Implicit associations, exemplified by LONGMEMEVAL's Multi-Session Reasoning tasks, require integrating semantically related evidence distributed across sessions and levels of abstraction; \texttt{MemTree} supports this setting through its multi-granular tree organization. 
Explicit associations, corresponding to LOCOMO's Multi-Hop questions, require multi-hop reasoning over structured paths connecting multiple memory units, which is supported by \texttt{MemGAS} through its multi-granularity graph organization.

For \textit{Temporal Reasoning}, the roadmap distinguishes two representative types of questions. 
Time retrieval, mainly represented by the temporal questions in LOCOMO, focuses on retrieving information associated with a specific temporal reference. For this type of questions, \texttt{MemOS} achieves the strongest performance. 
Time comparison, largely represented by LONGMEMEVAL's temporal reasoning tasks, requires comparing temporal relations among multiple pieces of information. As discussed in Section~8.2 of our revised paper (Exp.3. Context Scalability Analysis), increasing context length introduces higher noise levels. Therefore, we further distinguish low-noise and high-noise settings. Under low-noise histories, \texttt{LightMem} performs best, while under high-noise histories, \texttt{Mem0$^g$} achieves more robust performance.

For \textit{Knowledge Update}, the principal trade-off is between update effectiveness and token efficiency. \texttt{MemTree} provides better update performance when substantial update cost is acceptable, whereas the lightweight update design of \texttt{LightMem} is more appropriate under tight token budgets.

}

\revise{
\section{Memory Construction Cost of Zep}
\label{sec:app:zep}

The missing score indicates that {\tt Zep} did not complete memory construction on LONGMEMEVAL within two days in our unified experimental environment. The ``fast'' performance reported in {\tt Zep}'s documentation\footnote{\url{https://www.getzep.com/}} refers to retrieval latency, which is separately evaluated in Table~\ref{tab:latency}, rather than index construction time.
In our experiments, {\tt Zep} exhibits an average retrieval latency of 536.8\,ms, which is higher than the sub-200\,ms latency reported in its documentation. This difference arises because, to ensure a unified and controlled experimental setting, we use the open-source implementation released for the {\tt Zep} paper rather than its production API, which has undergone multiple subsequent iterations and optimizations. Regarding index construction, {\tt Zep} requires several hours to build the index for LOCOMO under our experimental setting. For LONGMEMEVAL, whose aggregate text volume is over $100\times$ that of LOCOMO, index construction could not be completed within two days.

To clarify this issue, we further analyze the sources of index construction overhead.
Specifically, {\tt Zep} represents each message as an episode and incrementally updates a temporal knowledge graph through {\tt Graphiti}\footnote{\url{https://github.com/getzep/graphiti}}. 
For each episode, it performs entity and fact extraction, entity resolution, fact deduplication, and conflict handling.
As the conversation grows, the expanding graph increases the candidate space for these operations, while longer context requires more information to be processed in LLM prompts. 
Consequently, later episodes generally incur higher processing costs, and repeatedly applying this pipeline over LONGMEMEVAL leads to substantial cumulative construction overhead.
}

\end{document}